\documentclass[10pt,twocolumn,letterpaper]{article}

\usepackage{cvpr}              % To produce the CAMERA-READY version
\usepackage{pifont}
\usepackage{bbding}
\usepackage{multirow}
\usepackage{amsmath}
\usepackage{graphicx}
\usepackage[dvipsnames]{xcolor}
\usepackage{color, colortbl}

\definecolor{LightCyan}{rgb}{0.88,1,1}
\definecolor{HighLight}{rgb}{0.96,0.92,0.96}
\DeclareMathOperator*{\argmax}{arg\,max}

\definecolor{cvprblue}{rgb}{0.21,0.49,0.74}
\usepackage[pagebackref,breaklinks,colorlinks,citecolor=cvprblue]{hyperref}

\def\paperID{4658} % *** Enter the Paper ID here
\def\confName{CVPR}
\def\confYear{2024}

\title{Dual Memory Networks:\\ A Versatile Adaptation Approach for Vision-Language Models}

\author{{Yabin Zhang$^{1,2}$ \quad Wenjie Zhu$^1$ \quad Hui Tang$^3$ \quad Zhiyuan Ma$^1$ \quad Kaiyang Zhou$^4$ \quad Lei Zhang$^{1,2,}$\thanks{Corresponding author.} } \\
	$^1$HKPolyU \qquad $^2$OPPO  \qquad $^3$HKUST \qquad $^4$HKBU \\
	{\tt\small \{csybzhang,cslzhang\}@comp.polyu.edu.hk}
}
\begin{document}
\maketitle

\begin{abstract}
	
With the emergence of pre-trained vision-language models like CLIP, how to adapt them to various downstream classification tasks has garnered significant attention in recent research. The adaptation strategies can be typically categorized into three paradigms: zero-shot adaptation, few-shot adaptation, and the recently-proposed training-free few-shot adaptation. Most existing approaches are tailored for a specific setting and can only cater to one or two of these paradigms. In this paper, we introduce a versatile adaptation approach that can effectively work under all three settings. 
Specifically, we propose the dual memory networks that comprise dynamic and static memory components. 
The static memory caches training data knowledge, enabling training-free few-shot adaptation, while the dynamic memory preserves historical test features online during the testing process, allowing for the exploration of additional data insights beyond the training set.
This novel capability enhances model performance in the few-shot setting and enables model usability in the absence of training data.
The two memory networks employ the same flexible memory interactive strategy, which can operate in a training-free mode and can be further enhanced by incorporating learnable projection layers. 
Our approach is tested across 11 datasets under the three task settings. Remarkably, in the zero-shot scenario, it outperforms existing methods by over 3\% and even shows superior results against methods utilizing external training data.
Additionally, our method exhibits robust performance against natural distribution shifts.
Codes are available at \url{https://github.com/YBZh/DMN}.
\end{abstract}

\section{Introduction}
\label{sec:intro}

Contrastive vision-language pre-training \cite{radford2021learning,jia2021scaling,li2021align,yang2022vision} has shown promising results in various downstream vision tasks, including 2D/3D perception \cite{zhou2022learning,zhang2022pointclip} and generation \cite{crowson2022vqgan,sanghi2023clip}. Among these models, CLIP \cite{radford2021learning} is arguably the most representative one due to its simplicity and effectiveness. Leveraging a vast collection of image-text pairs from the Internet, CLIP  aligns features across modalities, leading to notable zero-shot classification capabilities. To further enhance its performance on downstream tasks, numerous adaptation strategies have emerged, primarily employing frozen CLIP encoders in zero-shot and few-shot settings.

\begin{figure}
	\includegraphics[width=0.95\linewidth]{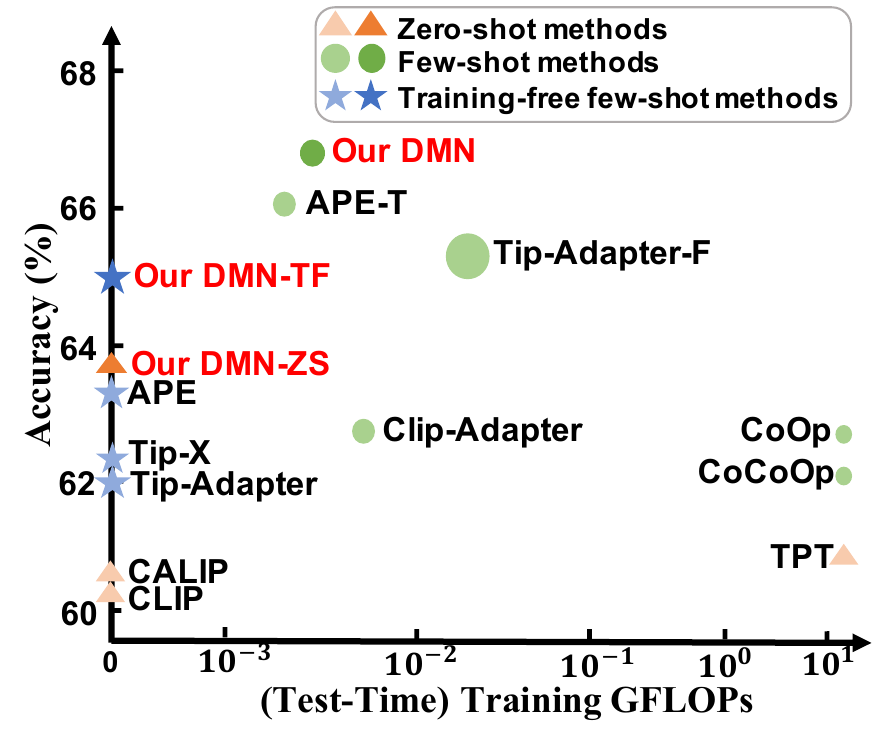}
	\vspace{-0.3cm}
	\caption{Illustration of the classification accuracy, (test-time) training GFLOPs, and learning parameters on zero-shot and 16-shot ImageNet classification. 
		The icon sizes denote the number of learnable parameters. 
		Our method is unique in its ability to work for all three task settings with superior results.
%		Our method is the only one that tackles all three task settings and achieves superior results in each of them.
	} \label{Fig:top_page}
\end{figure}

\begin{table*}
	\centering
	\begin{tabular}{l|c|ccc}
		\toprule
		 \multirow{2}{*}{Methods} & \multirow{2}{*}{\rotatebox{0}{No External Training Data}}  & \multicolumn{3}{c}{Task Settings} \\
		\cline{3-5}
		& & \rotatebox{0}{Zero-shot} & \rotatebox{0}{Few-shot} & \rotatebox{0}{TF Few-shot} \\
		\midrule
		TPT \cite{shu2022test} and \cite{zhou2023distribution,pratt2023does,menon2022visual,novack2023chils,ren2023chatgpt}    &  \Checkmark &  \Checkmark  & \textcolor{gray}{\XSolidBrush} & \textcolor{gray}{\XSolidBrush}  \\
		DiffTPT \cite{feng2023diverse}    & \textcolor{gray}{\XSolidBrush}  &  \Checkmark  & \textcolor{gray}{\XSolidBrush} & \textcolor{gray}{\XSolidBrush}  \\
		CoOp \cite{zhou2022learning} and \cite{zhou2022conditional,khattak2023maple,wortsman2022robust,zang2022unified,chen2022prompt,Shi_2023_ICCV,xing2023dual,zhu2023prompt,lu2022prompt,yu2023task} & \Checkmark &   \textcolor{gray}{\XSolidBrush} & \Checkmark&  \textcolor{gray}{\XSolidBrush} \\
		Tip-Adapter \cite{zhang2021tip}, and \cite{zhu2023not} & \Checkmark &  \textcolor{gray}{\XSolidBrush} & \Checkmark &  \Checkmark  \\
		SuS-X \cite{udandarao2023sus} &  \textcolor{gray}{\XSolidBrush} &  \textcolor{gray}{\XSolidBrush} & \Checkmark &  \Checkmark  \\
		CALIP \cite{guo2023calip}         &  \Checkmark &  \Checkmark&  \Checkmark&  \textcolor{gray}{\XSolidBrush}  \\
		CaFo \cite{zhang2023prompt} & \textcolor{gray}{\XSolidBrush} & \Checkmark &   \Checkmark &  \Checkmark   \\
		\midrule
		\rowcolor{HighLight} \textbf{DMN (Ours)}  &  \Checkmark &   \Checkmark &  \Checkmark&  \Checkmark\\
		\bottomrule
	\end{tabular}
\vspace{-0.1cm}
	\caption{Summary of adaptation methods for vision-language models. `Zero-shot', `Few-shot', and `TF Few-shot' represent the zero-shot adaptation, few-shot adaptation, and the recently introduced training-free few-shot adaptation, respectively. `No External Training Data' indicates that the approach does not utilize any synthetic training images from generation models or retrieved images via class names.   } \label{Tab:summary_clip_adapt}
		\vspace{-0.2cm}
\end{table*}

Most existing approaches are tailored for one specific task setting. Specifically, enhanced zero-shot performance is achieved by exploring additional insights from the test sample itself \cite{guo2023calip,shu2022test} or via enhanced text prompts \cite{pratt2023does,menon2022visual}.
In the few-shot setting, researchers typically insert adaptive parameters (\eg, Prompt \cite{zhou2022learning,khattak2023maple}, Adapter \cite{gao2023clip}, and Residual \cite{yu2023task}) into the pre-trained vision-language models and optimize these parameters using labeled training data. 
Recently, a training-free variant of few-shot adaptation has been proposed for resource-constrained applications \cite{zhang2021tip}. In this setting, no parameters are needed to learn, and thus much computational resources are saved.
While numerous methods have been introduced, they typically cater to only one or two task settings, as summarized in Tab. \ref{Tab:summary_clip_adapt}, thereby limiting their applicability.

In this work, we propose a versatile adaptation approach that works effectively for all the three task settings, as shown in Fig. \ref{Fig:top_page}.
Specifically, we propose the dual memory networks comprising dynamic and static memory components, producing sample-adaptive classifiers for each test point. 
The static memory network caches features of training data, generating the adaptive classifier for each test sample by adaptively weighting cached training features and thus enabling training-free few-shot adaptation. 
In contrast, the dynamic memory network preserves features of historical test samples during the testing process, introducing another adaptive classifier by adaptively weighting cached test features.
This allows us to explore additional data insights beyond the training samples, further enhancing the model's performance in the few-shot setting and extending its applications to the zero-shot setting where training data is absent.
These two types of memory networks employ the same memory interactive strategy, which is highly flexible. 
This strategy can be used in a training-free mode for zero-shot and training-free few-shot adaptations. In addition, it can be further enhanced by incorporating learnable projection layers in the traditional few-shot setting.

%These two types of memory networks employ the same memory interactive strategy, which is highly flexible. It can be either used in a training-free mode for zero-shot and training-free few-shot adaptations or be enhanced by incorporating learnable projection layers for the traditional few-shot setting

We evaluate our approach on 11 datasets. In particular, in the setting where external training data are unavailable, our method surpasses existing zero-shot methods by a significant margin of over 3\% by leveraging knowledge of historical test samples.
Even in comparison to methods that utilize external training data, our model still exhibits substantial advantages, outperforming the recent CaFo \cite{zhang2023prompt} by 1.48\%. 
These results highlight the crucial significance of historical test samples in the adaptation process, which is neglected in existing works.
It is worth emphasizing the efficiency of incorporating historical test knowledge with the dynamic memory network, as the memory interaction process involves only a single attention module.
Through the utilization of historical test knowledge, labeled training data, and vanilla text information, our approach significantly enhances few-shot performance, establishing a new state-of-the-art in both the few-shot and training-free few-shot settings.
Moreover, our method demonstrates excellent generalization capabilities to natural distribution shifts.
We summarize our contributions as follows:

\begin{itemize}
	\item We introduce a versatile adaptation strategy for pre-trained vision-language models, termed Dual Memory Networks (DMN), aimming to effectively address the tasks of zero-shot adaptation, few-shot adaptation, and training-free few-shot adaptation. To the best of our knowledge, \textit{this is the first work to enhance vision-language model adaptation across the three settings without the use of external training data}.
	\item DMN comprises static and dynamic memory networks that gather information from labeled training data and historical test data, respectively. The two memory networks employ a flexible interactive strategy, which can operate in a training-free mode and can be further enhanced with learnable projection layers.
	\item Our approach has been validated on 11 datasets with three task settings. In the zero-shot setting, it outperforms competitors by over 3\% and even surpasses methods using external training data. It also demonstrates robust performance against natural distribution shifts.
\end{itemize}

\section{Related Work}
\textbf{Adaptation of Vision-Language Models.}
Foundation models \cite{radford2021learning,kirillov2023segment,li2024univs,rombach2022high} have attracted increasing attention in downstream tasks recently \cite{sun2023improving,zhang2023improving,wu2024seesr,lin2023sam,li2023opensd}.
Pre-trained on vast collections of image-text pairs, vision-language models like CLIP exhibit remarkable zero-shot generalization capabilities across a range of downstream datasets \cite{radford2021learning}. 
Building upon CLIP, numerous methods have been introduced to adapt it to various downstream classification tasks, especially under the zero-shot and few-shot settings as summarized in Tab. \ref{Tab:summary_clip_adapt}.
In the zero-shot setting where labeled training data are unavailable, one primary research direction is how to extract richer information from the test samples \cite{guo2023calip,shu2022test,feng2023diverse,zhou2023distribution} and class names \cite{udandarao2023sus,feng2023diverse,novack2023chils,pratt2023does,menon2022visual}.
For the former group, CALIP \cite{guo2023calip} enhances feature extraction through attention mechanisms, and instance-adaptive prompts are explored using consistency regularization in \cite{shu2022test,feng2023diverse}.
Leveraging class names, some approaches \cite{zhang2023prompt,udandarao2023sus} generate synthetic training samples utilizing additional image generation models \cite{rombach2022high,dayma2021dall}, and others \cite{pratt2023does,menon2022visual,ren2023chatgpt} craft advanced text prompts by querying pre-trained large language models.

%DN \cite{zhou2023distribution} enhances the dot product classification using feature normalization with estimated mean representation of interested distribution. 
%When the training image is not available, how to make the full of class names also attracts attention. For example, augmented training images can be generated with the pre-trained generation models \cite{udandarao2023sus,feng2023diverse}. CHiLS \cite{novack2023chils} explore more knowledge from name of subclasses via the semantic hierarchy. 

To further unlock the potential of pre-trained CLIP models for downstream tasks, how to adapt the frozen CLIP model with a limited amount of labeled training data has attracted increasing attention, leading to the few-shot adaptation. Inspired by the parameter-efficient transfer learning \cite{lester2021power,houlsby2019parameter}, many methods propose to tune the pre-trained CLIP models with carefully designed prompts \cite{zhou2022learning,zhou2022conditional,chen2022prompt,khattak2023maple,zang2022unified,Khattak_2023_ICCV,chen2022prompt} and adapters \cite{gao2023clip}.
%\textcolor{red}{Splitting these prompts methods into subdirections.}
Besides, Lin \emph{et al.} \cite{lin2023multimodality}, Wortsman \emph{et al.} \cite{wortsman2022robust}, and Yu \emph{et al.} \cite{yu2023task} respectively investigate the cross-modal adaptation, weight ensembles, and task residuals for better CLIP adaptation.  
Recently, a training-free variant of few-shot adaptation has been proposed for resource-constrained applications \cite{zhang2021tip}, where computationally intensive model training is prohibited. Specifically, Tip-Adapter \cite{zhang2021tip} is a pioneering training-free few-shot approach, which caches the encoded features and labels of training images as task priors. Predictions are then derived based on the similarity between the test feature and cached features. Tip-Adapter is subsequently augmented with the integration of calibrated intra-modal distance as described in \cite{udandarao2023sus}, and through adaptive channel prior refinement as elaborated in \cite{zhu2023not}.
These training-free adaptation methods can be enhanced with optional model optimization by either tuning the cached features \cite{zhang2021tip} or adding learnable category residuals \cite{zhu2023not}.

Most aforementioned methods are tailored for a specific task setting and can only cater to one or two of these adaptation paradigms, as summarized in Tab. \ref{Tab:summary_clip_adapt}. 
Although existing few-shot methods can be applied to the zero-shot task by utilizing external training data through generation or searching \cite{zhang2023prompt,udandarao2023sus}, they may not fully meet the practical requirements of zero-shot applications, such as efficient and rapid adaptation to new tasks.
In contrast, we propose a versatile adaptation approach that can effectively handle all the three tasks without relying on any external training data. This is achieved by fully utilizing the training data and historical test samples via the proposed DMN framework, leading to the new state-of-the-art across all three adaptation settings.

\textbf{Memory Networks.}
Memory networks were initially introduced in the realm of Natural Language Processing. Inspired by the knowledge accumulation and recalling in human brain \cite{baddeley2000episodic,stokes2015activity}, they introduce an external memory component, allowing the storage and retrieval of historical knowledge to facilitate decision making \cite{weston2014memory,sukhbaatar2015end}. Subsequently, the concept of interactive memory, facilitating the storage and retrieval of historical information, has been adopted in various vision tasks, including classification \cite{karunaratne2021robust,santoro2016meta}, segmentation \cite{oh2019video,xie2021few,li2023mdqe}, and detection \cite{deng2019object,chen2020memory,li2022dual,li2023one}.
Recently, ideas reminiscent of memory networks have been introduced into CLIP adaptation \cite{zhang2021tip,udandarao2023sus}. However, the memory modules employed in their approaches are typically read-only and do not support real-time writing, akin to the static memory in our method. As expected, these approaches are unable to leverage historical test samples, limiting their performance in few-shot adaptation and impeding their application in zero-shot adaptation.
Our method stands out as the first to introduce a dynamic memory that supports both reading and writing operations for test data, while optionally maintaining a static memory for training data. By exploring all available data sources, our method can effectively handle all the three adaptation tasks and achieve superior performance.

\section{Method}

We first present a flexible memory interactive strategy for both dynamic and static memory networks. Then, we present these memory networks in detail.

\subsection{A Flexible Memory Interactive Strategy} \label{Subsec:general_mn}
Memory networks \cite{weston2014memory,sukhbaatar2015end} provide an effective mechanism to explicitly accumulate and recall knowledge, empowering better performance by utilizing the relevant historical information. 
A memory network typically comprises the following four abstract steps:
\begin{enumerate}
	\item Convert a new input $\mathbf{x}$ into the feature space.
	\item Update the memory $\mathbf{M}$ with $\mathbf{x}$.
	\item Read out an output given $\mathbf{x}$ and the current memory.
	\item Convert the output into the desired response. 
\end{enumerate}
In the following, we demonstrate how to instantiate these steps in CLIP adaptation, where the memory interaction strategy in steps 2 and 3 is our main focus.

\begin{figure*}
	\centering
	\includegraphics[width=0.87\linewidth]{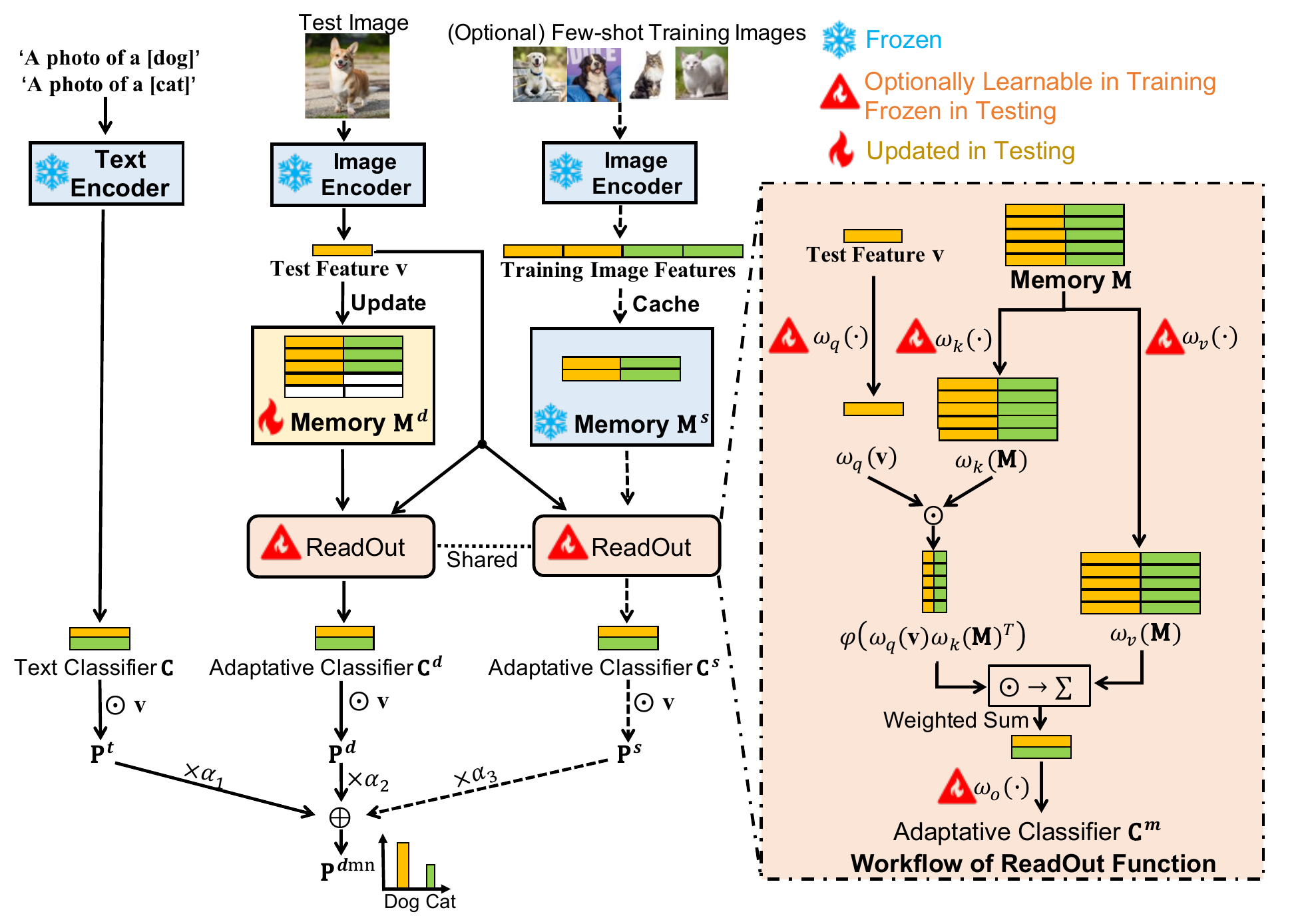}
		\vspace{-0.2cm}
	\caption{An illustration of the overall framework of our Dual Memory Networks (DMN), which integrates knowledge from three sources (\ie, text input, historical test data, and optional training images) to tackle the three types of adaptation tasks (\ie, zero-shot, few-shot, and the recently-proposed training-free few-shot adaptations).  } \label{Fig:framework}
		\vspace{-0.2cm}
\end{figure*}

We first present how to use CLIP to classify a test sample under the zero-shot setting. 
For a test image $\mathbf{x}$ within a downstream task of $C$ classes, we extract the visual representation $\mathbf{v} \in \mathbb{R}^{D}$ and textual representation $\mathbf{C} \in \mathbb{R}^{C\times D}$ with pre-trained CLIP encoders, where $D$ is the feature dimension. 
Both $\mathbf{v}$ and $\mathbf{C}$ are $L_2$ normalized along the $D$ dimension.
Then, the zero-shot prediction probability can be achieved by using text features $\mathbf{C}$ as the classifier:
\begin{equation} \label{Equ:text_pred}
\mathbf{P}^t = \mathrm{Softmax}\left(\mathbf{v}\mathbf{C}^{\top}\right) \in \mathbb{R}^{C},
\end{equation}
where the scaling parameter is omitted for simplicity. 

To instantiate the memory networks in CLIP adaption, it is natural to adopt the pretrained image encoders of CLIP to transform the input $\mathbf{x}$ to the image feature $\mathbf{v}$. 
We construct a category-split memory $\mathbf{M} \in \mathbb{R}^{C\times L \times D}$, where $L$ is the memory length for each category.
To update the memory $\mathbf{M}$ with $\mathbf{v}$, we simply store $\mathbf{v}$ in a `slot' of $\mathbf{M}$. Specifically, given the (pseudo) label $y \in [1,C]$ of the input image,  we locate the sub-memory $\mathbf{M}_y \in \mathbb{R}^{L \times D}$ corresponding to the category $y$, find an empty slot of it, say the $i^{th}$ row, denoted by $\mathbf{M}_{y,i}\in \mathbb{R}^{D}$, and update the memory as:
\begin{equation} \label{Eq:memory_write}
\mathbf{M}_{y,i} = \mathbf{v}.
\end{equation}
Besides the image feature, we also cache the corresponding prediction entropy estimated from $\mathbf{P}^t$, which is used to locate the slot to update when $\mathbf{M}_y$ is full. 
Specifically, if all rows of $\mathbf{M}_y$ are occupied by image features, we replace the row of maximum entropy in $\mathbf{M}_y$ with $\mathbf{v}$ if $\mathbf{v}$ exhibits smaller prediction entropy.
In other words, we store samples with lower prediction entropy in the memory.

Given the updated memory $\mathbf{M}$ and the test feature $\mathbf{v}$, we read out a sample adaptive classifier $\mathbf{C}^{m} \in \mathbb{R}^{C\times D}$ via cross-attention as:
\begin{equation} \label{Equ:general_memory_classifier_readout}
\mathbf{C}^{m}  = \textrm{ReadOut}(\mathbf{v}, \mathbf{M}),
\end{equation}
where the $y^{th}$ row of $\mathbf{C}^{m}$ is produced by using $\mathbf{v}$ as query and adopting memory $\mathbf{M}_y$ as key and value:
\begin{align} \label{Equ:general_memory_classifier}
\mathbf{C}^{m}_y = \omega_o \left( \varphi\left(\omega_q(\mathbf{v})\omega_k(\mathbf{M}_y)^{\top}\right)\omega_v(\mathbf{M}_y) \right).
\end{align}
The $\omega_q$, $\omega_k$, $\omega_v$ and $\omega_o$ respectively represent the project functions for query, key, value, and the output,  $\omega_q(\mathbf{v})\omega_k(\mathbf{M}_y)^{\top} \in \mathbb{R}^{1\times L}$ measures the cosine similarities between normalized features of $\omega_q(\mathbf{v})$ and $\omega_k(\mathbf{M}_y)$, and $\varphi(x) = \exp(-\beta (1-x))$ modulates the sharpness of $x$ with hyper-parameter $\beta$. 
%The adopted $\varphi(\cdot)$ achieves better results than the popular softmax function as analyzed in Sec. \ref{Subsec:ablation_analyses}.
Intuitively, $\mathbf{C}^{m}_y$ is the weighted combination of image features in $\mathbf{M}_y$, where the weight is based on the cosine similarity between test and memoried image features.
In other words, the sample adaptive classifier $\mathbf{C}^m$ is produced by image features, instead of the text features that produce the text classifier $\mathbf{C}$.

Finally, we follow \cref{Equ:text_pred} to convert the memory output $\mathbf{C}^{m}$ to the desired classification prediction, leading to the final memory response:
\begin{equation} \label{Equ:general_mn_pred}
\mathbf{P}^m = \mathrm{M2P}(\mathbf{v}, \mathbf{C}^m) =  \mathrm{Softmax}\left(\mathbf{v}\mathbf{C}^{m\top}\right)  \in \mathbb{R}^{C}.
\end{equation}
The $\mathbf{P}^m$ is the classification probability of test feature $\mathbf{v}$ with the  sample adaptive classifier $\mathbf{C}^m$.

The versatility of our memory interactive strategy across various task settings stems from the flexibility of the projection layer. 
Specifically, we define the projection function $\omega_*$ (covering $\omega_q, \omega_k, \omega_v$, and $\omega_o$) using a residual architecture:
\begin{equation} \label{Equ:projection_layer}
\omega_*(x) = \mathrm{L_2}\left(x + \mathrm{Linear}(x)\right),
\end{equation}
where $\mathrm{Linear}(\cdot)$ represents a linear layer with all parameters initialized to zero and $\mathrm{L_2}(\cdot)$ indicates the $L_2$ normalization along feature dimension. In the training-free setting, the projection function $\omega_*(\cdot)$ degenerates to $\omega_*(x) = x$, given the $L_2$ normalized input $x$.
Therefore, the memory interaction is conducted in the vanilla feature space of CLIP.  
Given labeled training samples, we can explore a more efficient feature space for memory interaction by optimizing the linear layers with the classification objective.
Next, we present the dynamic and static memory networks based on this flexible interactive strategy.

\subsection{Dynamic Memory Network} \label{subsec:dmn}
The dynamic memory networks accumulate historical test samples in the test process and is activated for all task settings. 
Firstly, we introduce a dynamic memory $\mathbf{M}^{d} \in \mathbb{R}^{C\times L \times D}$ initialized with zero values.
Given the test feature $\mathbf{v}$, we update the memory $\mathbf{M}^{d}$ using Eq. (\ref{Eq:memory_write}) with the estimated pseudo label $y$ from the text classifier:
\begin{equation} \label{Equ:pseudo_labeling_dmn}
y = \argmax_j \mathbf{P}^t_j.
\end{equation}
Given the updated memory $\mathbf{M}^{d}$ and the test feature $\mathbf{v}$, we can read out a sample adaptive classifier $\mathbf{C}^{d}$ with the readout function in Eq. (\ref{Equ:general_memory_classifier_readout}) as:
\begin{equation} \label{Equ:dynamic_memory_readout}
\mathbf{C}^{d}  = \textrm{ReadOut}(\mathbf{v}, \widehat{\mathbf{M}}^{d}),
\end{equation}
where $\widehat{\mathbf{M}}^{d} = [\mathbf{M}^{d}, \mathbf{C} ] \in \mathbb{R}^{C\times (L+1)\times D}$ is the extended memory with text feature. 
Such a memory extension actually initializes the  $\mathbf{C}^{d}$ with the text classifier $\mathbf{C}$, considering that the memory $\mathbf{M}^{d}$ is initialized with zero values.
As more image features are written into the memory, the classifier $\mathbf{C}^{d}$ is gradually refined with cached image features, utilizing the historical test samples in the testing process. Finally, the sample classification probability with the dynamic memory network is introduced with Eq. (\ref{Equ:general_mn_pred}) as:
\begin{equation}
\mathbf{P}^d = \mathrm{M2P}(\mathbf{v}, \mathbf{C}^d)  \in \mathbb{R}^{C}.
\end{equation}
The prediction $\mathbf{P}^d$ utilizes knowledge of historical test samples, including the current one, whose effectiveness is analyzed in Sec. \ref{Subsec:ablation_analyses}.

\subsection{Dual Memory Networks} \label{subsec:fmn}
In this section, we present the full version of our versatile DMN, which comprises the aforementioned dynamic memory network and the following static memory network. The overall framework is shown in Fig. \ref{Fig:framework}.
For a $C$-way-$K$-shot task with $K$ training images per category, one may opt to utilize these samples by extending the dynamic memories with image features of these data, \ie, updating $\widehat{\mathbf{M}}^{d} = [\mathbf{M}^{d}, \mathbf{M}^{s}, \mathbf{C} ] \in \mathbb{R}^{C\times (L+K+1)\times D}$ in Eq. (\ref{Equ:dynamic_memory_readout}), where $\mathbf{M}^{s} \in \mathbb{R}^{C\times K\times D}$ is the aggregation of image features of $CK$ training samples. 
Although this simple strategy brings certain improvement, we argue that the valuable knowledge from labeled data may gradually get diluted as the dynamic memory fills up.
This dilution results in a degraded performance (see Fig. \ref{Fig:dynamic_vs_static} for more analyses).

To make full use of labeled data, we additionally maintain one static memory, \ie, $\mathbf{M}^{s}$, and introduce another sample adaptive classifier using these labeled data only. 
As described by its name, the static memory $\mathbf{M}^{s}$ keeps unchanged after creation.
Given the static memory $\mathbf{M}^{s}$ and the test feature $\mathbf{v}$, we can read out a sample adaptive classifier $\mathbf{C}^{s}$ with the readout function in Eq. (\ref{Equ:general_memory_classifier_readout}) as:
\begin{equation} \label{Equ:static_memory_readout}
\mathbf{C}^{s}  = \textrm{ReadOut}(\mathbf{v}, \mathbf{M}^{s}).
\end{equation}
The corresponding prediction probability is:
\begin{equation}
\mathbf{P}^s = \mathrm{M2P}(\mathbf{v}, \mathbf{C}^s)  \in \mathbb{R}^{C}.
\end{equation}
The prediction $\mathbf{P}^s$ is based on the knowledge of labeled training data, which are complement to the text knowledge in $\mathbf{P}^t$ and historical test knowledge in $\mathbf{P}^d$. The final prediction is obtained by aggregating the three knowledge sources:
\begin{equation} \label{Equ:dmn_three_pred}
\mathbf{P}^{dmn} = \alpha_1 \mathbf{P}^{t}  +  \alpha_2 \mathbf{P}^{d} + \alpha_3 \mathbf{P}^{s},
\end{equation}
where $\alpha_{1\sim 3}$ denote the weights for text prediction, prediction of dynamic memory network, and prediction of static memory network, respectively.

\begin{table} \small
	\centering
	\begin{tabular}{l|l|ccc}
		\toprule
		 Variants & Adaptation Settings &  $\mathbf{M}^{d}$ & $\mathbf{M}^{s}$ & $\omega_*$\\
		\midrule
		DMN-ZS & Zero-shot              &  \Checkmark & \textcolor{gray}{\XSolidBrush}  & \textcolor{gray}{\XSolidBrush}  \\
		DMN-TF & Training-free Few-shot  &  \Checkmark & \Checkmark  & \textcolor{gray}{\XSolidBrush}  \\
		DMN      & Few-shot  &  \Checkmark  & \Checkmark  & \Checkmark \\
		\bottomrule
	\end{tabular}
\vspace{-0.2cm}
	\caption{Summary of our DMN variants for different adaptation tasks. The `$\mathbf{M}^{d}$' and `$\mathbf{M}^{s}$' respectively represent whether the dynamic and the static memory networks are activated and `$\omega_*$' indicates whether the projection layers are optimized. } \label{Tab:variant_summary}
	\vspace{-0.2cm}
\end{table}

\begin{table*} \footnotesize
	\centering
	\begin{tabular}{l|ccccccccccc|c}
		\toprule
		 Method & ImageNet
		& Flower & DTD & Pets & Cars & UCF & Caltech & Food & SUN & Aircraft & EuroSAT & Mean   \\
		\midrule
		CLIP-{RN50} \cite{radford2021learning} & 58.16 & 61.75 & 40.37 & 83.57 & 55.70 & 58.84 & 85.88 & 73.97 & 58.80 & 15.66 & 23.69 & 56.04 \\
		%Emsemble & 61.67 & 65.98 & 48.46 & 85.01 & 57.29 & 63.28 & 89.53 & 76.09 & 63.09 & 19.53 & 38.01 & 60.63 \\  %% tip+cupl
		DN \cite{zhou2023distribution}  & 60.16  & 63.32  & 41.21 & 81.92  & 56.55 & 55.60 & 87.25 & 74.64 & 59.11 & 17.43 & 28.31 & 56.86 \\
		%		Ensemble &  & 62.77 & 40.37 & 82.97 & 55.89 & 59.48 & 87.26 & 74.82 & 60.85 & 16.11 & 25.79 & 56.63 \\ 80 prompts
		TPT \cite{shu2022test} & 60.74 &  62.69 & 40.84 & 84.49 & 58.46 & 60.82 & 87.02 & 74.88 & 61.46 & 17.58 & 28.33 & 57.94 \\
		VisDesc \cite{menon2022visual} &  59.68 &  65.37  & 41.96 & 82.39 & 54.76 & 58.47 & 88.11 & 76.80 & 59.84 & 16.26 & 37.60 & 58.29\\
		Ensemble \cite{zhang2021tip} & 60.32 &  66.10 & 40.07 & 85.83 & 55.71 & 61.33 & 83.94 & 77.32 & 58.53 & 17.10 & 37.54 & 58.53 \\  %% tip-prompt.
		CALIP \cite{guo2023calip} & 60.57 &  66.38 &  42.39 & 86.21 & 56.27 & 61.72 & 87.71 & 77.42 &58.59 & 17.76 & 38.90 & 59.45 \\
		DiffTPT \cite{feng2023diverse}$^*$ & 60.80 & 63.53 & 40.72 & 83.40 & 60.71 & 62.67 & 86.89 & \textbf{79.21} & 62.72 & 17.60 & 41.04 & 59.94 \\
		%+ DM-img  & 60.37 & 42.08 & & & 61.46 & & 69.97 & & 17.46  \\
		%+ DM-both & 62.20 & 42.73 & 85.36 & 60.15 & 63.12 & 87.30 & 74.79 & 62.47 & 18.42 & 29.26 & 58.58 \\
		CuPL \cite{pratt2023does}     &  61.45  & 65.44  & 48.64 & 84.84 & 57.28 & 58.97 & 89.29 & 76.94 & 62.55 & 19.59 & 38.38 & 60.31 \\
		SuS-X-SD-C \cite{udandarao2023sus}$^*$ &  61.84  & 67.72  & 50.59 & 85.34 & 57.27 & 61.54 & 89.53 & 77.58 & 62.95 & 19.47 & 45.57 & 61.76\\
		%\textbf{Ours} & 63.54 & 43.03 & 84.98 & 60.29 & 62.20 & 89.25 & 75.88 & 63.46 & 18.99 & 31.68 & \textbf{59.33} \\  %%% prompt, a photo of a
		%% 66.02   42.91   85.72   56.20   61.72   87.91   75.77   61.16   17.01   36.22   590.64  %% baseline, a photo of a
		CaFo \cite{zhang2023prompt}$^*$  & 62.74 & 66.54 & 50.24 & \textbf{87.49} & 58.45 & 63.67 & \textbf{90.91} & 77.53 & 63.16 & 21.06 & 42.73 & 62.23 \\
			\rowcolor{HighLight}  \textbf{DMN-ZS (Ours)} & \textbf{63.87} & \textbf{67.93} & \textbf{50.41} & 86.78 & \textbf{60.02} & \textbf{65.34} & 90.14 & 76.70 & \textbf{64.39} & \textbf{22.77} & \textbf{48.72} & \textbf{63.71}  \\
		\midrule
		CLIP-{ViTB/16} \cite{radford2021learning} & 66.73 & 67.44 & 44.27 & 88.25 & 65.48 & 65.13 & 93.35 & 83.65 & 62.59 & 23.67 & 42.01 & 63.87 \\
		%+ DM-both   &  70.73 & 49.47 & 89.10 & 68.28 & 70.29 & 93.91 & 84.21 & 67.95 & 24.87 & 53.41 &  67.22 \\
		Ensemble \cite{zhang2021tip} & 68.34 &  66.99 & 45.04 & 86.92 & 66.11 & 65.16 & 93.55 & 82.86 & 65.63 & 23.22 & 50.42 & 64.93 \\
		TPT \cite{shu2022test} & 68.98& 68.98 & 47.75 & 87.79 & 66.87 & 68.04 & 94.16 & 84.67 & 65.50 & 24.78 & 42.44 & 65.45 \\
		%+ DM-both  & 69.96 & 49.53 & 88.06 & 68.36 & 70.31 & 93.91 & 84.87 & 68.07 & 25.26 & 52.68 \\
		DiffTPT \cite{feng2023diverse}$^*$ & 70.30 & 70.10 & 47.00 & 88.20 & 67.01 & 68.22 & 92.49 & \textbf{87.23} & 65.74 & 25.60 & 43.13 & 65.91 \\
		%		\textbf{Ours} & 69.91 & 49.82 & 89.07 & 70.03 & 71.03 & 94.85 & 84.56 & 69.54 & 24.96 & 61.65 & \textbf{68.54} \\  %% prompt, a photo of a
		%		70.73   44.09   89.04   66.20   67.46   93.91   84.28   66.25   24.81   48.35   655.12, baseline, tip prompt.
			\rowcolor{HighLight}  \textbf{DMN-ZS (Ours)} & \textbf{72.25} & \textbf{74.49}  & \textbf{55.85} & \textbf{92.04} & \textbf{67.96} & \textbf{72.51} & \textbf{95.38} & 85.08 & \textbf{70.18} & \textbf{30.03} & \textbf{59.43} & \textbf{70.72} \\
		\bottomrule
	\end{tabular}
\vspace{-0.2cm}
	\caption{Zero-shot classification performance on eleven downstream datasets, where results with $^*$ are achieved with external training data. } \label{Tab:zero_shot}
\end{table*}
\begin{figure*}[ht]
	\centering
	\begin{subfigure}{0.33\textwidth}
		\includegraphics[width=\textwidth]{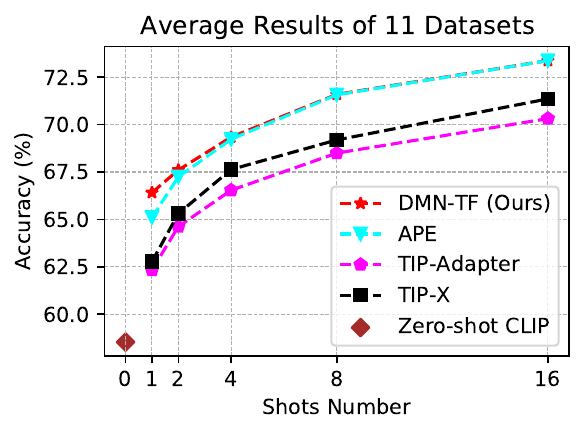}
	\end{subfigure}
	\hfill
	\begin{subfigure}{0.33\textwidth}
		\includegraphics[width=\textwidth]{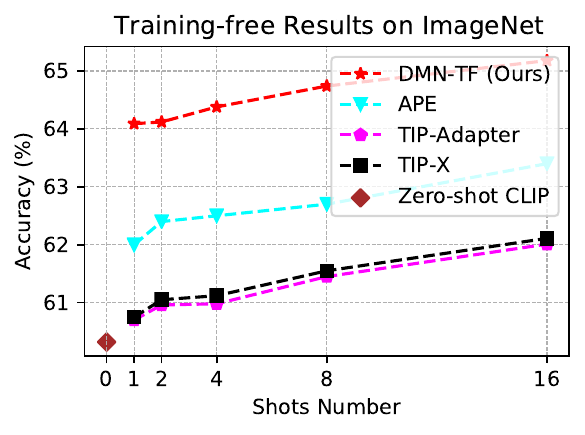}
	\end{subfigure}
	\hfill
	\begin{subfigure}{0.33\textwidth}
		\includegraphics[width=\textwidth]{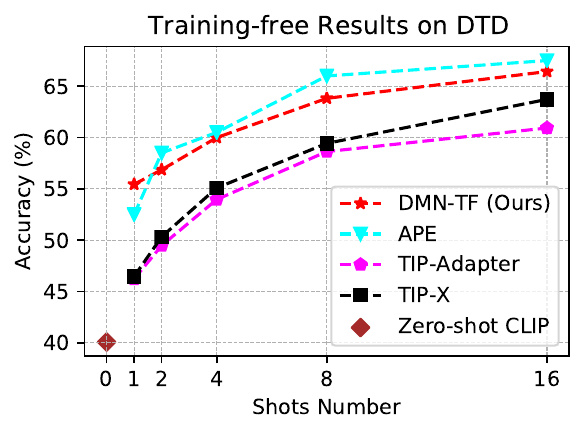}
	\end{subfigure}
	\vspace{-0.3cm}
	\caption{Training-free few-shot results with a ResNet50 backbone. Full results on $11$ classification datasets are presented in Fig. \ref{Fig:full_tf_res50}. } \label{Fig:training_free_dsmn}
	\vspace{-0.2cm}
\end{figure*}

Our DMN is a versatile adaptation approach for vision-language models that handles three task settings, \ie, zero-shot, few-shot, and training-free few-shot adaptations. 
Considering the inherent variations among different task settings, the implementation of our DMN exhibits subtle differences.
For example, in the training-free setting, such as zero-shot and the training-free few-shot adaptations, we adopt the initialized projection layers in Eq. (\ref{Equ:projection_layer}) and conduct memory interaction in the vanilla CLIP feature space, while we finetune these projection layers and explore more efficient feature space for the traditional few-shot setting. 
To distinguish our results under different task settings, we term the DMN variants with respect to zero-shot, few-shot, and training-free few-shot settings as DMN-ZS, DMN, and DMN-TF, respectively. We summarize these variants in Tab. \ref{Tab:variant_summary}.

\begin{figure*}[ht]
	\centering
	\begin{subfigure}{0.3\textwidth}
		\includegraphics[width=\textwidth]{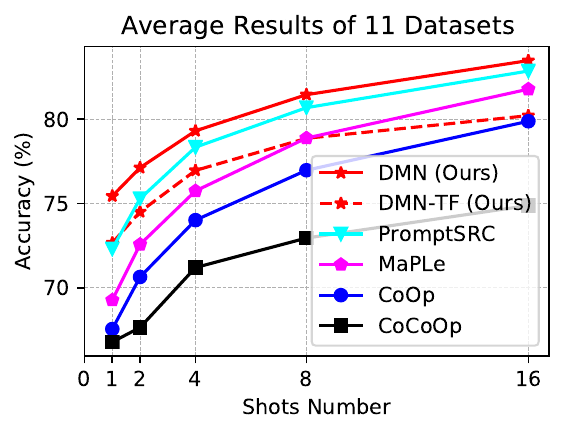}
	\end{subfigure}
	\hfill
	\begin{subfigure}{0.3\textwidth}
		\includegraphics[width=\textwidth]{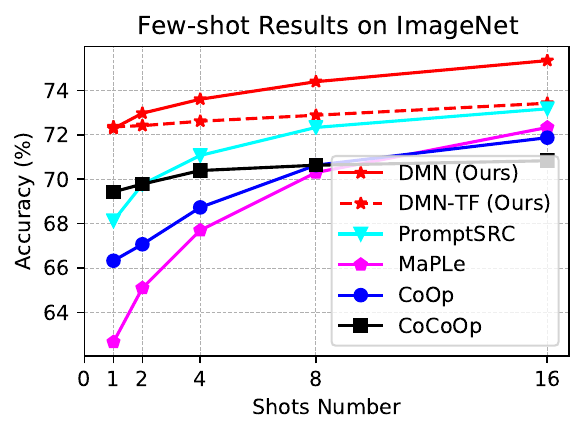}
	\end{subfigure}
	\hfill
	\begin{subfigure}{0.3\textwidth}
		\includegraphics[width=\textwidth]{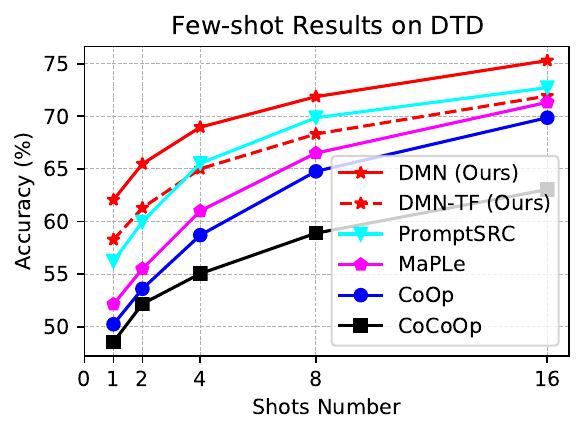}
	\end{subfigure}
	\vspace{-0.3cm}
	\caption{Few-shot performance with ViTB/16 backbone, where the full results on $11$ classification datasets are presented in Fig. \ref{Fig:full_tr_vit}. } \label{Fig:few_shot_vit}
\end{figure*}

\begin{figure*}[ht]
	\centering
	\begin{subfigure}{0.3\textwidth}
		\includegraphics[width=\textwidth]{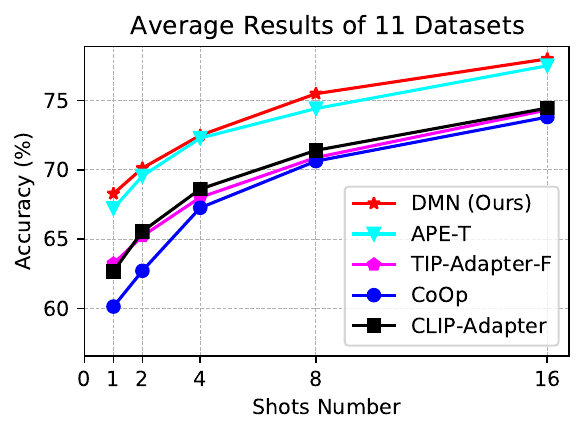}
		%		\caption{}
	\end{subfigure}
	\hfill
	\begin{subfigure}{0.3\textwidth}
		\includegraphics[width=\textwidth]{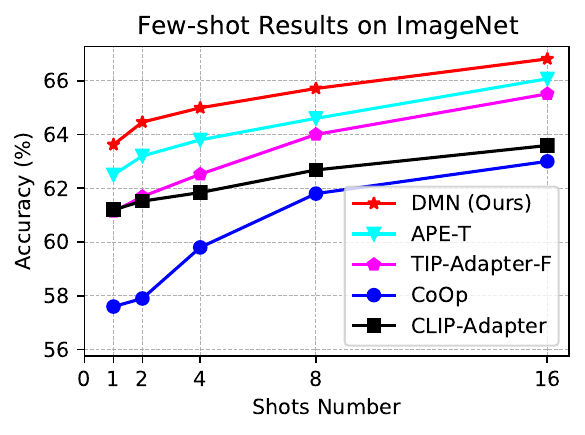}
		%		\caption{}
	\end{subfigure}
	\hfill
	\begin{subfigure}{0.3\textwidth}
		\includegraphics[width=\textwidth]{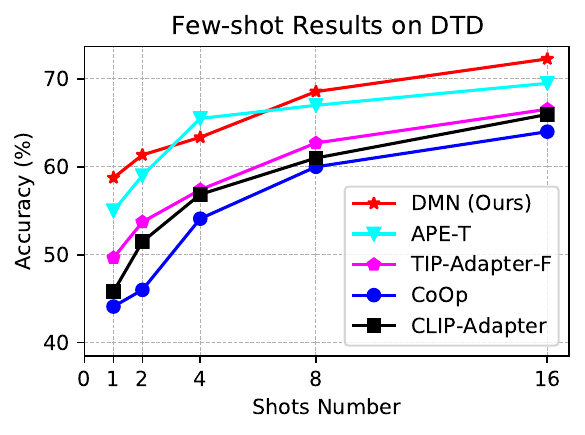}
		%		\caption{}
	\end{subfigure}
	\vspace{-0.3cm}
	\caption{Few-shot performance with ResNet50 backbone, where the full results on $11$ classification datasets are presented in Fig. \ref{Fig:full_tr_res50}.   }   \label{Fig:few_shot_tr_res50}
	\vspace{-0.cm}
\end{figure*}

\section{Experiments}
% We first present the experiment settings. Then, we evaluate our method and conduct careful ablation studies on standard benchmark datasets.

\subsection{Experiment Settings}
\textbf{Datasets.} We validate our method on 11 classification benchmarks, including ImageNet \cite{deng2009imagenet}, Flowers102 \cite{nilsback2008automated},  DTD \cite{cimpoi2014describing}, OxfordPets \cite{parkhi2012cats}, StandfordCars \cite{krause20133d}, UCF101 \cite{soomro2012ucf101}, Caltech101 \cite{fei2004learning}, Food101 \cite{bossard2014food}, SUN397 \cite{xiao2010sun}, FGVCAircraft \cite{maji2013fine}, and EuroSAT \cite{helber2019eurosat}. We also evaluate the robustness of DMN to natural distribution shifts \cite{zhang2020unsupervised,zhang2022exact} on four ImageNet variants, \ie, ImageNet-V2 \cite{recht2019imagenet}, ImageNet-A \cite{hendrycks2021nae}, ImageNet-R \cite{hendrycks2021many}, and ImageNet-Sketch \cite{wang2019learning}. 
%Besides 2D image datasets, we also validate our method on 3D point cloud datasets of ModelNet10 \cite{xx}, ModelNet40 \cite{xx}, and ScanObjectNN \cite{xx}.s

\textbf{Settings.} We adopt visual encoders of ResNet50 \cite{he2016deep} and VIT-B/16 \cite{dosovitskiy2020image} pretrained by CLIP. We follow existing works to conduct the image split in few-shot learning and adopt the textual prompt in \cite{zhang2021tip,pratt2023does}. 
Inspired by \cite{shu2022test}, we enhance the robust pseudo label estimation in Eq. \ref{Equ:pseudo_labeling_dmn} with view augmentation and confidence selection.
We search the optimal prediction weights, \ie, $\alpha_{1\sim 3}$, for each downstream task, while illustrate that the fixed weights generalize well within each task setting. 
We train the DMN with AdamW optimizer \cite{loshchilov2017decoupled}, where we adopt the cosine annealing learning schedule with the initial learning rate of 1e-4 and set the batch size as 128.  
We train the model for 20 epochs for most datasets except for the Flower102 and EuroSAT, where 100 epochs are adopted.

\begin{table} \footnotesize
	\begin{tabular}{l|c|cccc}
		\toprule
		 Method & ImageNet & -A & -V2 & -R & -Sketch  \\
		\midrule
		CLIP-RN50 \cite{radford2021learning} & 58.16 & 21.83 & 51.41 & 56.15 & 33.37  \\
		Ensemble & 59.81 & 23.24 & 52.91 & 60.72 & 35.48  \\
		%+ DM & 60.63 & 27.83 & 54.19 & \textcolor{red}{53.80} & 35.77 & 46.44 &  42.90 \\
		TPT \cite{shu2022test}   & 60.74 & 26.67 & 54.70 & 59.11 & 35.09  \\
		%+ DW & 61.42 & 28.76 & 54.30 & \textcolor{red}{56.28} & 36.35 & 47.42 & 43.92 \\
		CALIP \cite{guo2023calip} & 60.57 & 23.96 & 53.70 & 60.81 & 35.61   \\
		DiffTPT \cite{feng2023diverse} & 60.80 &  \textbf{31.06} & 55.80 & 58.80 & 37.10   \\
		CoCoOp$^*$ \cite{zhou2022conditional} & 62.81 & 23.32 & 55.72 & 57.74 & 34.48  \\
		CoOp$^*$ \cite{zhou2022learning} & 63.33 & 23.06 & 55.40 & 56.60 & 34.67  \\
		%+ DM & 64.83 & 29.36 & 58.40 & 55.01 & 36.76 \\
		%%% with 7 ensemble text prompt.
		%	\textbf{Ours}   & \textbf{62.58} & 29.77 & 56.18 & 61.09 & 38.04 & \textbf{49.53} & \textbf{46.27} \\  %% prompt, a photo of a
		%	60.33   23.76   53.39   60.57   35.50   233.55  %%%% baseline, tip prompt.
		\rowcolor{HighLight} \textbf{DMN-ZS (Ours)} & \textbf{63.87} & 28.57 &  \textbf{56.12}  & \textbf{61.44}  & \textbf{39.84}   \\
		\midrule
		CLIP-ViT-B/16 \cite{radford2021learning} & 66.73 & 47.87 & 60.86 & 73.98 & 46.09  \\
		%+ DM-image   & 68.53 & 57.32 & 63.99 & 70.94 & 48.13 \\
		%+ DM-both     & 69.89 & 57.87 & 64.24 & 75.19 & 49.65 \\
		Ensemble & 68.34 & 49.89 & 61.88 & 77.65 & 48.24  \\
		TPT \cite{shu2022test} & 68.98 & 54.77 & 63.45 & 77.06 & 47.94  \\
		%+ DM-image & 68.55 & 57.35 & 63.99 & 70.93 & 48.14\\
		%+ DM-both   & 70.32 & 58.87 & 64.32 & \textcolor{red}{76.85} & 49.87 \\
		DiffTPT \cite{feng2023diverse} & 70.30 & 55.68 & 65.10 & 75.00 & 46.80   \\
		MaPLe$^*$ \cite{khattak2023maple}  & 70.72 & 50.90 & 64.07 & 76.98 & 49.15  \\
		CoCoOp$^*$ \cite{zhou2022conditional} & 71.02 & 50.63 & 64.07 & 76.18 & 48.75  \\
		CoOp$^*$ \cite{zhou2022learning} & 71.51 & 49.71 & 64.20 & 75.21 & 47.99  \\
		PromptSRC$^*$ \cite{Khattak_2023_ICCV} & 71.27 & 50.90 & 64.35 & 77.80 & 49.55  \\
		%	\textbf{Ours} & \textbf{71.57} & 59.60 & 65.56 & 78.74 & 51.66 & \textbf{65.43} & \textbf{63.89} \\  %%%% prompt, a photo of a
		\rowcolor{HighLight} \textbf{DMN-ZS (Ours)}  & \textbf{72.25} &   \textbf{58.28}  &  \textbf{65.17} &  \textbf{78.55} &  \textbf{53.20}  \\
		\bottomrule
	\end{tabular}
\vspace{-0.2cm}
	\caption{Robustness to Natural Distribution Shifts. Results with $^*$ are tuned on ImageNet using 16-shot training samples per category, while other methods do not require labeled training data.  } \label{Tab:ood_class}
	\vspace{-0.1cm}
\end{table}

\subsection{Performance Evaluation} \label{Sec:performance_results}
\textbf{Zero-shot DMN-ZS Results.}
We first present the experimental results under the zero-shot adaptation setting, where the significance of historical test knowledge becomes particularly pronounced. 
As illustrated in Tab. \ref{Tab:zero_shot}, our method surpasses its closest competitors that do not involve external training data, such as CALIP and TPT. Specifically, we observe improvements of 3.40\% and 5.27\% when employing ResNet-50 and ViTB/16 backbones, respectively. 
Compared to approaches like TPT \cite{shu2022test}, which necessitate model optimization on test samples, the memory interactions within our DMN do not introduce any test time optimization, substantially accelerating the inference speed, as shown in Tab. \ref{Tab:computation_efficiency}. 

To tackle the zero-shot challenge, some approaches utilize labeled synthetic training samples generated from pre-trained image generation models \cite{udandarao2023sus,zhang2023prompt}.
By treating these synthetic labeled data like genuine labeled data, the zero-shot problem can be tackled through few-shot approaches.
While these strategies offer notable performance gains, the generation of synthetic data and subsequent model optimization come with considerable computational overheads, failing to meet the efficient adaptation requirement in zero-shot setting. 
In contrast, incorporating historical test knowledge with our dynamic memory network is considerably faster.
Interestingly, even when compared to techniques that employ synthetic training data, our approach maintains a distinct advantage, highlighting the superiority of historical test samples over synthetic training data.

%To address the zero-shot challenge, certain approaches resort to the introduction of labeled synthetic training samples derived from pre-trained image generation models. By treating these synthetic labeled data akin to genuine labeled data, the zero-shot problem can be tackled through few-shot paradigms. 
%While such strategies manifest significant performance gains, the generation of synthetic data and subsequent model optimization incur considerable computational overheads. Interestingly, even compared with techniques that employ synthetic training data, our approach maintains a distinct advantage, underlining the superiority of historical test samples over synthetic training data. 

\textbf{Training-free Few-shot DMN-TF Results.}
We compare our DMN-TF with the training-free few-shot methods of Tip-Adapter \cite{zhang2021tip}, Tip-X \cite{udandarao2023sus}, and the recent APE \cite{zhu2023not}. 
As illustrated in Fig. \ref{Fig:training_free_dsmn}, our method achieves a superior advantage with one training sample per category. The advantage gradually diminishes with additional training samples. 
% Note that the APE approach enhances Tip-Adapter by highlighting discriminative feature channels, which can be integrated into our method for boosted performance.   

\begin{figure*}
	\centering
	\begin{subfigure}{0.25\linewidth}
		\includegraphics[width=\linewidth]{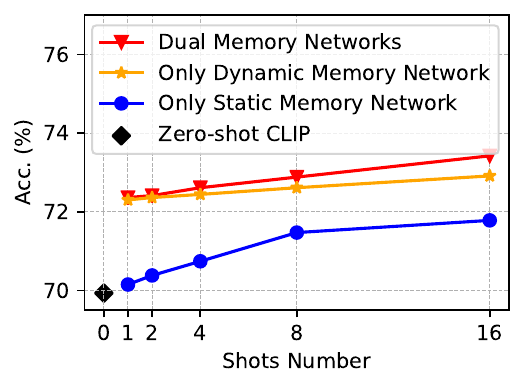}
		\caption{Two memory networks}
		\label{Fig:dynamic_vs_static}
	\end{subfigure}
	\hfill
	\begin{subfigure}{0.24\linewidth}
		\includegraphics[width=\linewidth]{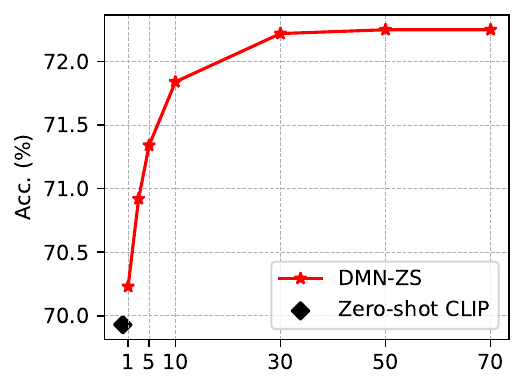}
		\caption{Memory length $L$}
		\label{Fig:memory_length}
	\end{subfigure}
	\hfill
	\begin{subfigure}{0.24\linewidth}
		\includegraphics[width=\linewidth]{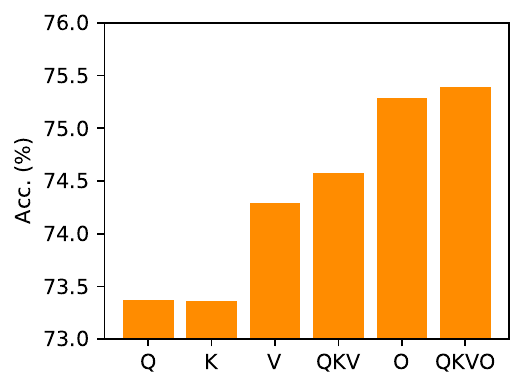}
		\caption{Position of projection layers}
		\label{Fig:proj_position}
	\end{subfigure}
	\hfill
	\begin{subfigure}{0.24\linewidth}
		\includegraphics[width=\linewidth]{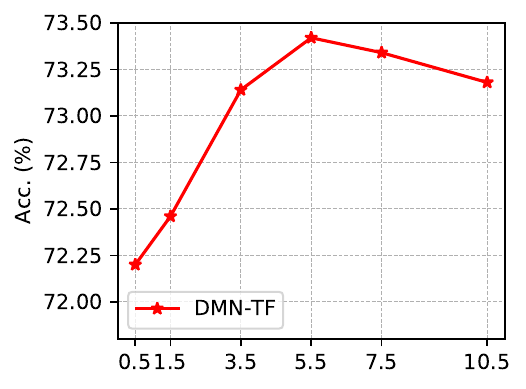}
		\caption{Values of $\beta$}
		\label{Fig:beta_value}
	\end{subfigure}
	\caption{Analyses on (a) static and dynamic memory networks, (b)  memory length of the dynamic memory, (b) position of projection layers, and (d) values of $\beta$ in Eq. (\ref{Equ:general_memory_classifier}).}
	\label{Fig:analyses_all}
	\vspace{-0.2cm}
\end{figure*}

\textbf{Few-shot DMN Results.}
We compare our method with seven few-shot adaptation methods of CoOp \cite{zhou2022learning}, CoCoOp \cite{zhou2022conditional}, MaPLe \cite{khattak2023maple}, PromptSRC \cite{Khattak_2023_ICCV}, CLIP-Adapter \cite{gao2023clip}, Tip-Adapter-F \cite{zhang2021tip}, and APE-T \cite{zhu2023not}.  All methods employed for comparison do not utilize external training data. As evidenced by the results averaged over eleven datasets shown in Fig. \ref{Fig:few_shot_vit} and Fig. \ref{Fig:few_shot_tr_res50}, our DMN consistently surpasses competing approaches, maintaining superiority with different backbone architectures and varying numbers of training samples. 
On individual datasets, although our method occasionally lags behind some competing methods in certain settings, it achieves consistent gains on the acknowledged ImageNet dataset, affirming its effectiveness.

\textbf{Generalization to Natural Distribution Shifts.}
As illustrated in Tab. \ref{Tab:ood_class}, our method not only achieves superior performance on traditional ImageNet dataset, but also generalizes well to the samples with natural distribution shifts, validating its robustness.

\subsection{Ablation and Analyses} \label{Subsec:ablation_analyses}
\textbf{Dynamic Memory Network vs. Static Memory Network.}
To analyze the roles of dynamic and static memory networks individually, we introduce two degenerated versions of DMN with dynamic or static memory network only. 
We illustrate the results under the training-free few-shot setting in Fig. \ref{Fig:dynamic_vs_static}.
Both dynamic and static memory networks significantly outperform zero-shot CLIP, and the larger the training sample size, the greater the improvement. Results with dynamic memory network surpass those with static memory network, confirming the importance of historical test samples. The optimal results are achieved by combining the advantages of both memory networks, validating their complementarity.

%As illustrated in Fig. \ref{Fig:memory_ablation}, the results of SMN and SMN-TF gradually improve as the number of training samples increases, validating the effectiveness of static memory network. The role of dynamic memory network can be observed through the significant advantage of DMN-ZS/DMN-TF/DMN over Zero-shot CLIP/SMN-TF/SMN. As expect, the improvement brought by the dynamic memory network gradually diminishes as the number of training samples increases. For instance, under the one-shot setting, the DMN-TF outperforms SMN-TF by 2.21\%, but this improvement gradually decreases to 1.64\% under the 16-shot setting.
%On the contrary, as the number of training samples increases, greater advantages are obtained by exploring an effective memory interactive space through the tuning of projection functions $\omega_*(\cdot)$. This is demonstrated by the increasing advantage of SMN over SMN-TF, from 0.06\% at 1-shot to 2.11\% at 16-shot settings.
%
%\textcolor{red}{Results with dynamic memory only.}

\textbf{Memory Length.}
As shown in Fig. \ref{Fig:memory_length}, the classification accuracy gradually increases as the memory length increases and saturates when the memory length exceeds 30. In all experiments, we set the memory length to 50.

\textbf{Position of Projection Layers.}
We report the results with different projection layers in Fig. \ref{Fig:proj_position}, where Q, K, V and O represent the $\omega_q$, $\omega_k$, $\omega_v$ and $\omega_o$, respectively.
We observe that all these projection layers bring improvement and the output projection, \ie, $\omega_o$, contributes the most to the results. 
We adopt the QKVO strategy in all experiments.

\textbf{Values of $\beta$.}
Results with different values of $\beta$ are illustrated in Fig. \ref{Fig:beta_value}. We set $\beta$=5.5 in all experiments.

%\begin{figure}[tb]
%	\centering
%	\includegraphics[width=0.6\linewidth]{images/memory_length.pdf}
%	\caption{Results with different lengths of dynamic memory.} \label{Fig:memory_length}
%\end{figure}
%\begin{figure}[tb]
%	\centering
%	\includegraphics[width=0.6\linewidth]{images/projection_position.pdf}
%	\caption{Analyses on position of projection layers.} \label{Fig:proj_position}
%\end{figure}

\textbf{Computation Efficiency.}
As summarized in Tab. \ref{Tab:computation_efficiency}, in zero-shot and training-free few-shot settings, our approach does not introduce any learning parameter, maintaining fast inference speed. In classical few-shot learning, our method achieves fast adaptation
% and new state-of-the-art results 
by introducing a small amount of training computation and learnable parameters.

\begin{table}[tb]\footnotesize
	\centering
	\begin{tabular}{l|cccc}
		\toprule
		\textbf{Methods} & \textbf{Train} & \textbf{Test} & \textbf{GFLOPs} & \textbf{Param.} \\
		\midrule
		\textcolor{gray}{Zero-shot}  \\
		CLIP \cite{radford2021learning}   & --  & 10.1ms & 0 & 0  \\
		CALIP \cite{guo2023calip}   & -- & 10.2ms & 0 & 0  \\
		TPT \cite{shu2022test}    & --  & 436ms            & $>$10& 0.01M    \\
		\rowcolor{HighLight}\textbf{DMN-ZS (Ours)} & -- & 10.7ms  & 0 & 0  \\
		\midrule
		\textcolor{gray}{Few-shot}  \\
		Tip-Adapter \cite{zhang2021tip}      & -- & 10.4ms & 0 & 0  \\
		%Tip-X            & -- & ?? & 0 & 0 & 62.11 \\
		APE  \cite{zhu2023not}            & --  & 10.4ms & 0 & 0  \\
		\rowcolor{HighLight}\textbf{DMN-TF (Ours)} & -- & 10.7ms & 0 & 0  \\
		\cline{2-5}
		CoOp \cite{zhou2022learning}          & 14 h & 10.2ms & $>$10 & 0.01M   \\
		CLIP-Adapter \cite{gao2023clip}  & 50 min & 10.4ms & 0.004 & 0.52M  \\
		Tip-Adapter-F \cite{zhang2021tip}  & 5 min & 10.4ms & 0.030 & 16.3M  \\
		APE-T \cite{zhu2023not}             & 5 min & 10.4ms & 0.002 & 0.51M  \\
		\rowcolor{HighLight} \textbf{DMN (Ours)} & 5 min & 10.7ms & 0.033 & 4.20M   \\
		\bottomrule
	\end{tabular}
	\caption{Analyses of computation efficiency on zero-shot and 16-shot ImageNet with a ResNet50 backbone. `Training' measures the training time, 
		`GFLOPs' are calculated during training or test-time training with gradient back-propagation, and `Param.' presents the number of learnable parameters. 
		% `Test' reports the inference speed, measured with a batch size of 32, except for TPT, which uses a batch size of 1 for testing.
%		`Test' reports the inference speed
		 Results are achieved with a NVIDIA RTX A6000 GPU. }
	\label{Tab:computation_efficiency}
\end{table}

Due to the limit of space, more analyses on classifier weights, non-linear function $\varphi(\cdot)$, and test data order can be found in the Supplementary Material.

%\begin{table}[htb]
%	\centering 
%	\begin{tabular}{c|ccccc}
%		\hline
%		Position & Q & K & V & Output   \\
%		\hline
%		Acc       & 73.29 & 73.46 & 74.43 & 75.39  \\
%		position & all     & all-shared & qkv     & qkv-shared \\
%		Acc        & 75.05& 74.70       & 74.40  & 74.37  \\
%		\hline
%	\end{tabular}
%	\caption{Caption}
%	\label{tab:my_label}
%\end{table}

\section{Conclusion}
In this paper, we proposed a versatile adaptation approach, named Dual Memory Networks (DMN), for vision-language models. By leveraging historical test data and few-shot training samples with dynamic and static memory networks, our DMN can handle all the three commonly used task settings: zero-shot, few-shot, and training-free few-shot adaptations, outperforming existing methods designed for single-task scenarios. 
Notably, the integration of the dynamic memory network, which utilizes historical test knowledge, distinguished our approach from previous research that overlooked this knowledge source. 
Nonetheless, our approach had some limitations due to the introduction of two external memories. For instance, in the case of 16-shot ImageNet adaptation, the dynamic and static memories occupied storage space of 204.8MB and 65.5MB, respectively. This may pose challenges for its applications to  storage-constrained scenarios.

%The key distinction of our approach from existing methods lied in the integration of the dynamic memory network, which utilized historical test knowledge to boost adaptation performance across various settings,
%This was an aspect that has been overlooked in previous research.

%\begin{figure}
%  \centering
%  \begin{subfigure}{0.49\linewidth}
%    \fbox{\rule{0pt}{2in} \rule{.9\linewidth}{0pt}}
%    \caption{An example of a subfigure.}
%    \label{fig:short-a}
%  \end{subfigure}
%  \hfill
%  \begin{subfigure}{0.49\linewidth}
%    \fbox{\rule{0pt}{2in} \rule{.9\linewidth}{0pt}}
%    \caption{Another example of a subfigure.}
%    \label{fig:short-b}
%  \end{subfigure}
%  \caption{Example of a short caption, which should be centered.}
%  \label{fig:short}
%\end{figure}

%\input{sec/2_formatting}
%\input{sec/3_finalcopy}
{
    \small
    \bibliographystyle{ieeenat_fullname}
    \bibliography{main}

\begin{thebibliography}{77}
\providecommand{\natexlab}[1]{#1}
\providecommand{\url}[1]{\texttt{#1}}
\expandafter\ifx\csname urlstyle\endcsname\relax
  \providecommand{\doi}[1]{doi: #1}\else
  \providecommand{\doi}{doi: \begingroup \urlstyle{rm}\Url}\fi

\bibitem[Baddeley(2000)]{baddeley2000episodic}
Alan Baddeley.
\newblock The episodic buffer: a new component of working memory?
\newblock \emph{Trends in cognitive sciences}, 4\penalty0 (11):\penalty0
  417--423, 2000.

\bibitem[Bossard et~al.(2014)Bossard, Guillaumin, and
  Van~Gool]{bossard2014food}
Lukas Bossard, Matthieu Guillaumin, and Luc Van~Gool.
\newblock Food-101--mining discriminative components with random forests.
\newblock In \emph{Computer Vision--ECCV 2014: 13th European Conference,
  Zurich, Switzerland, September 6-12, 2014, Proceedings, Part VI 13}, pages
  446--461. Springer, 2014.

\bibitem[Chen et~al.(2022)Chen, Yao, Song, Li, Rao, and Zhang]{chen2022prompt}
Guangyi Chen, Weiran Yao, Xiangchen Song, Xinyue Li, Yongming Rao, and Kun
  Zhang.
\newblock Prompt learning with optimal transport for vision-language models.
\newblock \emph{arXiv preprint arXiv:2210.01253}, 2022.

\bibitem[Chen et~al.(2020)Chen, Cao, Hu, and Wang]{chen2020memory}
Yihong Chen, Yue Cao, Han Hu, and Liwei Wang.
\newblock Memory enhanced global-local aggregation for video object detection.
\newblock In \emph{Proceedings of the IEEE/CVF conference on computer vision
  and pattern recognition}, pages 10337--10346, 2020.

\bibitem[Cimpoi et~al.(2014)Cimpoi, Maji, Kokkinos, Mohamed, and
  Vedaldi]{cimpoi2014describing}
Mircea Cimpoi, Subhransu Maji, Iasonas Kokkinos, Sammy Mohamed, and Andrea
  Vedaldi.
\newblock Describing textures in the wild.
\newblock In \emph{Proceedings of the IEEE conference on computer vision and
  pattern recognition}, pages 3606--3613, 2014.

\bibitem[Crowson et~al.(2022)Crowson, Biderman, Kornis, Stander, Hallahan,
  Castricato, and Raff]{crowson2022vqgan}
Katherine Crowson, Stella Biderman, Daniel Kornis, Dashiell Stander, Eric
  Hallahan, Louis Castricato, and Edward Raff.
\newblock Vqgan-clip: Open domain image generation and editing with natural
  language guidance.
\newblock In \emph{European Conference on Computer Vision}, pages 88--105.
  Springer, 2022.

\bibitem[Dayma et~al.(2021)Dayma, Patil, Cuenca, Saifullah, Abraham, Le~Khac,
  Melas, and Ghosh]{dayma2021dall}
Boris Dayma, Suraj Patil, Pedro Cuenca, Khalid Saifullah, Tanishq Abraham,
  Ph{\'u}c Le~Khac, Luke Melas, and Ritobrata Ghosh.
\newblock Dall{\textperiodcentered} e mini.
\newblock \emph{HuggingFace. com. https://huggingface.
  co/spaces/dallemini/dalle-mini (accessed Sep. 29, 2022)}, 2021.

\bibitem[Deng et~al.(2019)Deng, Hua, Song, Zhang, Xue, Ma, Robertson, and
  Guan]{deng2019object}
Hanming Deng, Yang Hua, Tao Song, Zongpu Zhang, Zhengui Xue, Ruhui Ma, Neil
  Robertson, and Haibing Guan.
\newblock Object guided external memory network for video object detection.
\newblock In \emph{Proceedings of the IEEE/CVF International Conference on
  Computer Vision}, pages 6678--6687, 2019.

\bibitem[Deng et~al.(2009)Deng, Dong, Socher, Li, Li, and
  Fei-Fei]{deng2009imagenet}
Jia Deng, Wei Dong, Richard Socher, Li-Jia Li, Kai Li, and Li Fei-Fei.
\newblock Imagenet: A large-scale hierarchical image database.
\newblock In \emph{2009 IEEE conference on computer vision and pattern
  recognition}, pages 248--255. Ieee, 2009.

\bibitem[Dosovitskiy et~al.(2020)Dosovitskiy, Beyer, Kolesnikov, Weissenborn,
  Zhai, Unterthiner, Dehghani, Minderer, Heigold, Gelly,
  et~al.]{dosovitskiy2020image}
Alexey Dosovitskiy, Lucas Beyer, Alexander Kolesnikov, Dirk Weissenborn,
  Xiaohua Zhai, Thomas Unterthiner, Mostafa Dehghani, Matthias Minderer, Georg
  Heigold, Sylvain Gelly, et~al.
\newblock An image is worth 16x16 words: Transformers for image recognition at
  scale.
\newblock \emph{arXiv preprint arXiv:2010.11929}, 2020.

\bibitem[Fei-Fei et~al.(2004)Fei-Fei, Fergus, and Perona]{fei2004learning}
Li Fei-Fei, Rob Fergus, and Pietro Perona.
\newblock Learning generative visual models from few training examples: An
  incremental bayesian approach tested on 101 object categories.
\newblock In \emph{2004 conference on computer vision and pattern recognition
  workshop}, pages 178--178. IEEE, 2004.

\bibitem[Feng et~al.(2023)Feng, Yu, Liu, Khan, and Zuo]{feng2023diverse}
Chun-Mei Feng, Kai Yu, Yong Liu, Salman Khan, and Wangmeng Zuo.
\newblock Diverse data augmentation with diffusions for effective test-time
  prompt tuning.
\newblock In \emph{Proceedings of the IEEE/CVF International Conference on
  Computer Vision}, pages 2704--2714, 2023.

\bibitem[Gao et~al.(2023)Gao, Geng, Zhang, Ma, Fang, Zhang, Li, and
  Qiao]{gao2023clip}
Peng Gao, Shijie Geng, Renrui Zhang, Teli Ma, Rongyao Fang, Yongfeng Zhang,
  Hongsheng Li, and Yu Qiao.
\newblock Clip-adapter: Better vision-language models with feature adapters.
\newblock \emph{International Journal of Computer Vision}, pages 1--15, 2023.

\bibitem[Guo et~al.(2023)Guo, Zhang, Qiu, Ma, Miao, He, and Cui]{guo2023calip}
Ziyu Guo, Renrui Zhang, Longtian Qiu, Xianzheng Ma, Xupeng Miao, Xuming He, and
  Bin Cui.
\newblock Calip: Zero-shot enhancement of clip with parameter-free attention.
\newblock In \emph{Proceedings of the AAAI Conference on Artificial
  Intelligence}, pages 746--754, 2023.

\bibitem[He et~al.(2016)He, Zhang, Ren, and Sun]{he2016deep}
Kaiming He, Xiangyu Zhang, Shaoqing Ren, and Jian Sun.
\newblock Deep residual learning for image recognition.
\newblock In \emph{Proceedings of the IEEE conference on computer vision and
  pattern recognition}, pages 770--778, 2016.

\bibitem[Helber et~al.(2019)Helber, Bischke, Dengel, and
  Borth]{helber2019eurosat}
Patrick Helber, Benjamin Bischke, Andreas Dengel, and Damian Borth.
\newblock Eurosat: A novel dataset and deep learning benchmark for land use and
  land cover classification.
\newblock \emph{IEEE Journal of Selected Topics in Applied Earth Observations
  and Remote Sensing}, 12\penalty0 (7):\penalty0 2217--2226, 2019.

\bibitem[Hendrycks et~al.(2021{\natexlab{a}})Hendrycks, Basart, Mu, Kadavath,
  Wang, Dorundo, Desai, Zhu, Parajuli, Guo, Song, Steinhardt, and
  Gilmer]{hendrycks2021many}
Dan Hendrycks, Steven Basart, Norman Mu, Saurav Kadavath, Frank Wang, Evan
  Dorundo, Rahul Desai, Tyler Zhu, Samyak Parajuli, Mike Guo, Dawn Song, Jacob
  Steinhardt, and Justin Gilmer.
\newblock The many faces of robustness: A critical analysis of
  out-of-distribution generalization.
\newblock \emph{ICCV}, 2021{\natexlab{a}}.

\bibitem[Hendrycks et~al.(2021{\natexlab{b}})Hendrycks, Zhao, Basart,
  Steinhardt, and Song]{hendrycks2021nae}
Dan Hendrycks, Kevin Zhao, Steven Basart, Jacob Steinhardt, and Dawn Song.
\newblock Natural adversarial examples.
\newblock \emph{CVPR}, 2021{\natexlab{b}}.

\bibitem[Houlsby et~al.(2019)Houlsby, Giurgiu, Jastrzebski, Morrone,
  De~Laroussilhe, Gesmundo, Attariyan, and Gelly]{houlsby2019parameter}
Neil Houlsby, Andrei Giurgiu, Stanislaw Jastrzebski, Bruna Morrone, Quentin
  De~Laroussilhe, Andrea Gesmundo, Mona Attariyan, and Sylvain Gelly.
\newblock Parameter-efficient transfer learning for nlp.
\newblock In \emph{International Conference on Machine Learning}, pages
  2790--2799. PMLR, 2019.

\bibitem[Jia et~al.(2021)Jia, Yang, Xia, Chen, Parekh, Pham, Le, Sung, Li, and
  Duerig]{jia2021scaling}
Chao Jia, Yinfei Yang, Ye Xia, Yi-Ting Chen, Zarana Parekh, Hieu Pham, Quoc Le,
  Yun-Hsuan Sung, Zhen Li, and Tom Duerig.
\newblock Scaling up visual and vision-language representation learning with
  noisy text supervision.
\newblock In \emph{International conference on machine learning}, pages
  4904--4916. PMLR, 2021.

\bibitem[Karunaratne et~al.(2021)Karunaratne, Schmuck, Le~Gallo, Cherubini,
  Benini, Sebastian, and Rahimi]{karunaratne2021robust}
Geethan Karunaratne, Manuel Schmuck, Manuel Le~Gallo, Giovanni Cherubini, Luca
  Benini, Abu Sebastian, and Abbas Rahimi.
\newblock Robust high-dimensional memory-augmented neural networks.
\newblock \emph{Nature communications}, 12\penalty0 (1):\penalty0 2468, 2021.

\bibitem[Khattak et~al.(2023{\natexlab{a}})Khattak, Rasheed, Maaz, Khan, and
  Khan]{khattak2023maple}
Muhammad~Uzair Khattak, Hanoona Rasheed, Muhammad Maaz, Salman Khan, and
  Fahad~Shahbaz Khan.
\newblock Maple: Multi-modal prompt learning.
\newblock In \emph{Proceedings of the IEEE/CVF Conference on Computer Vision
  and Pattern Recognition}, pages 19113--19122, 2023{\natexlab{a}}.

\bibitem[Khattak et~al.(2023{\natexlab{b}})Khattak, Wasim, Naseer, Khan, Yang,
  and Khan]{Khattak_2023_ICCV}
Muhammad~Uzair Khattak, Syed~Talal Wasim, Muzammal Naseer, Salman Khan,
  Ming-Hsuan Yang, and Fahad~Shahbaz Khan.
\newblock Self-regulating prompts: Foundational model adaptation without
  forgetting.
\newblock In \emph{Proceedings of the IEEE/CVF International Conference on
  Computer Vision (ICCV)}, pages 15190--15200, 2023{\natexlab{b}}.

\bibitem[Kirillov et~al.(2023)Kirillov, Mintun, Ravi, Mao, Rolland, Gustafson,
  Xiao, Whitehead, Berg, Lo, et~al.]{kirillov2023segment}
Alexander Kirillov, Eric Mintun, Nikhila Ravi, Hanzi Mao, Chloe Rolland, Laura
  Gustafson, Tete Xiao, Spencer Whitehead, Alexander~C Berg, Wan-Yen Lo, et~al.
\newblock Segment anything.
\newblock In \emph{Proceedings of the IEEE/CVF International Conference on
  Computer Vision}, pages 4015--4026, 2023.

\bibitem[Krause et~al.(2013)Krause, Stark, Deng, and Fei-Fei]{krause20133d}
Jonathan Krause, Michael Stark, Jia Deng, and Li Fei-Fei.
\newblock 3d object representations for fine-grained categorization.
\newblock In \emph{Proceedings of the IEEE international conference on computer
  vision workshops}, pages 554--561, 2013.

\bibitem[Lester et~al.(2021)Lester, Al-Rfou, and Constant]{lester2021power}
Brian Lester, Rami Al-Rfou, and Noah Constant.
\newblock The power of scale for parameter-efficient prompt tuning.
\newblock \emph{arXiv preprint arXiv:2104.08691}, 2021.

\bibitem[Li et~al.(2021)Li, Selvaraju, Gotmare, Joty, Xiong, and
  Hoi]{li2021align}
Junnan Li, Ramprasaath Selvaraju, Akhilesh Gotmare, Shafiq Joty, Caiming Xiong,
  and Steven Chu~Hong Hoi.
\newblock Align before fuse: Vision and language representation learning with
  momentum distillation.
\newblock \emph{Advances in neural information processing systems},
  34:\penalty0 9694--9705, 2021.

\bibitem[Li et~al.(2023{\natexlab{a}})Li, Li, Xiang, and Zhang]{li2023mdqe}
Minghan Li, Shuai Li, Wangmeng Xiang, and Lei Zhang.
\newblock Mdqe: Mining discriminative query embeddings to segment occluded
  instances on challenging videos.
\newblock In \emph{Proceedings of the IEEE/CVF conference on computer vision
  and pattern recognition}, pages 10524--10533, 2023{\natexlab{a}}.

\bibitem[Li et~al.(2024)Li, Li, Zhang, and Zhang]{li2024univs}
Minghan Li, Shuai Li, Xindong Zhang, and Lei Zhang.
\newblock Univs: Unified and universal video segmentation with prompts as
  queries.
\newblock \emph{arXiv preprint arXiv:2402.18115}, 2024.

\bibitem[Li et~al.(2022)Li, He, Li, and Zhang]{li2022dual}
Shuai Li, Chenhang He, Ruihuang Li, and Lei Zhang.
\newblock A dual weighting label assignment scheme for object detection.
\newblock In \emph{Proceedings of the IEEE/CVF conference on computer vision
  and pattern recognition}, pages 9387--9396, 2022.

\bibitem[Li et~al.(2023{\natexlab{b}})Li, Li, Li, He, and Zhang]{li2023one}
Shuai Li, Minghan Li, Ruihuang Li, Chenhang He, and Lei Zhang.
\newblock One-to-few label assignment for end-to-end dense detection.
\newblock In \emph{Proceedings of the IEEE/CVF conference on computer vision
  and pattern recognition}, pages 7350--7359, 2023{\natexlab{b}}.

\bibitem[Li et~al.(2023{\natexlab{c}})Li, Li, Wang, and Zhang]{li2023opensd}
Shuai Li, Minghan Li, Pengfei Wang, and Lei Zhang.
\newblock Opensd: Unified open-vocabulary segmentation and detection.
\newblock \emph{arXiv preprint arXiv:2312.06703}, 2023{\natexlab{c}}.

\bibitem[Lin et~al.(2023{\natexlab{a}})Lin, Liu, Lu, and Jia]{lin2023sam}
Jiehong Lin, Lihua Liu, Dekun Lu, and Kui Jia.
\newblock Sam-6d: Segment anything model meets zero-shot 6d object pose
  estimation.
\newblock \emph{arXiv preprint arXiv:2311.15707}, 2023{\natexlab{a}}.

\bibitem[Lin et~al.(2023{\natexlab{b}})Lin, Yu, Kuang, Pathak, and
  Ramanan]{lin2023multimodality}
Zhiqiu Lin, Samuel Yu, Zhiyi Kuang, Deepak Pathak, and Deva Ramanan.
\newblock Multimodality helps unimodality: Cross-modal few-shot learning with
  multimodal models.
\newblock In \emph{Proceedings of the IEEE/CVF Conference on Computer Vision
  and Pattern Recognition}, pages 19325--19337, 2023{\natexlab{b}}.

\bibitem[Loshchilov and Hutter(2017)]{loshchilov2017decoupled}
Ilya Loshchilov and Frank Hutter.
\newblock Decoupled weight decay regularization.
\newblock \emph{arXiv preprint arXiv:1711.05101}, 2017.

\bibitem[Lu et~al.(2022)Lu, Liu, Zhang, Liu, and Tian]{lu2022prompt}
Yuning Lu, Jianzhuang Liu, Yonggang Zhang, Yajing Liu, and Xinmei Tian.
\newblock Prompt distribution learning.
\newblock In \emph{Proceedings of the IEEE/CVF Conference on Computer Vision
  and Pattern Recognition}, pages 5206--5215, 2022.

\bibitem[Maji et~al.(2013)Maji, Rahtu, Kannala, Blaschko, and
  Vedaldi]{maji2013fine}
Subhransu Maji, Esa Rahtu, Juho Kannala, Matthew Blaschko, and Andrea Vedaldi.
\newblock Fine-grained visual classification of aircraft.
\newblock \emph{arXiv preprint arXiv:1306.5151}, 2013.

\bibitem[Menon and Vondrick(2022)]{menon2022visual}
Sachit Menon and Carl Vondrick.
\newblock Visual classification via description from large language models.
\newblock \emph{arXiv preprint arXiv:2210.07183}, 2022.

\bibitem[Nilsback and Zisserman(2008)]{nilsback2008automated}
Maria-Elena Nilsback and Andrew Zisserman.
\newblock Automated flower classification over a large number of classes.
\newblock In \emph{2008 Sixth Indian conference on computer vision, graphics \&
  image processing}, pages 722--729. IEEE, 2008.

\bibitem[Novack et~al.(2023)Novack, McAuley, Lipton, and Garg]{novack2023chils}
Zachary Novack, Julian McAuley, Zachary~Chase Lipton, and Saurabh Garg.
\newblock Chils: Zero-shot image classification with hierarchical label sets.
\newblock In \emph{International Conference on Machine Learning}, pages
  26342--26362. PMLR, 2023.

\bibitem[Oh et~al.(2019)Oh, Lee, Xu, and Kim]{oh2019video}
Seoung~Wug Oh, Joon-Young Lee, Ning Xu, and Seon~Joo Kim.
\newblock Video object segmentation using space-time memory networks.
\newblock In \emph{Proceedings of the IEEE/CVF International Conference on
  Computer Vision}, pages 9226--9235, 2019.

\bibitem[Parkhi et~al.(2012)Parkhi, Vedaldi, Zisserman, and
  Jawahar]{parkhi2012cats}
Omkar~M Parkhi, Andrea Vedaldi, Andrew Zisserman, and CV Jawahar.
\newblock Cats and dogs.
\newblock In \emph{2012 IEEE conference on computer vision and pattern
  recognition}, pages 3498--3505. IEEE, 2012.

\bibitem[Pratt et~al.(2023)Pratt, Covert, Liu, and Farhadi]{pratt2023does}
Sarah Pratt, Ian Covert, Rosanne Liu, and Ali Farhadi.
\newblock What does a platypus look like? generating customized prompts for
  zero-shot image classification.
\newblock In \emph{Proceedings of the IEEE/CVF International Conference on
  Computer Vision}, pages 15691--15701, 2023.

\bibitem[Radford et~al.(2021)Radford, Kim, Hallacy, Ramesh, Goh, Agarwal,
  Sastry, Askell, Mishkin, Clark, et~al.]{radford2021learning}
Alec Radford, Jong~Wook Kim, Chris Hallacy, Aditya Ramesh, Gabriel Goh,
  Sandhini Agarwal, Girish Sastry, Amanda Askell, Pamela Mishkin, Jack Clark,
  et~al.
\newblock Learning transferable visual models from natural language
  supervision.
\newblock In \emph{International conference on machine learning}, pages
  8748--8763. PMLR, 2021.

\bibitem[Recht et~al.(2019)Recht, Roelofs, Schmidt, and
  Shankar]{recht2019imagenet}
Benjamin Recht, Rebecca Roelofs, Ludwig Schmidt, and Vaishaal Shankar.
\newblock Do imagenet classifiers generalize to imagenet?
\newblock In \emph{International conference on machine learning}, pages
  5389--5400. PMLR, 2019.

\bibitem[Ren et~al.(2023)Ren, Su, and Liu]{ren2023chatgpt}
Zhiyuan Ren, Yiyang Su, and xiaoming Liu.
\newblock Chatgpt-powered hierarchical comparisons for image classification.
\newblock \emph{Advances in neural information processing systems}, 2023.

\bibitem[Rombach et~al.(2022)Rombach, Blattmann, Lorenz, Esser, and
  Ommer]{rombach2022high}
Robin Rombach, Andreas Blattmann, Dominik Lorenz, Patrick Esser, and Bj{\"o}rn
  Ommer.
\newblock High-resolution image synthesis with latent diffusion models.
\newblock In \emph{Proceedings of the IEEE/CVF conference on computer vision
  and pattern recognition}, pages 10684--10695, 2022.

\bibitem[Sanghi et~al.(2023)Sanghi, Fu, Liu, Willis, Shayani, Khasahmadi,
  Sridhar, and Ritchie]{sanghi2023clip}
Aditya Sanghi, Rao Fu, Vivian Liu, Karl~DD Willis, Hooman Shayani, Amir~H
  Khasahmadi, Srinath Sridhar, and Daniel Ritchie.
\newblock Clip-sculptor: Zero-shot generation of high-fidelity and diverse
  shapes from natural language.
\newblock In \emph{Proceedings of the IEEE/CVF Conference on Computer Vision
  and Pattern Recognition}, pages 18339--18348, 2023.

\bibitem[Santoro et~al.(2016)Santoro, Bartunov, Botvinick, Wierstra, and
  Lillicrap]{santoro2016meta}
Adam Santoro, Sergey Bartunov, Matthew Botvinick, Daan Wierstra, and Timothy
  Lillicrap.
\newblock Meta-learning with memory-augmented neural networks.
\newblock In \emph{International conference on machine learning}, pages
  1842--1850. PMLR, 2016.

\bibitem[Shi and Yang(2023)]{Shi_2023_ICCV}
Cheng Shi and Sibei Yang.
\newblock Logoprompt:synthetic text images can be good visual prompts for
  vision-language models.
\newblock In \emph{Proceedings of the IEEE/CVF International Conference on
  Computer Vision (ICCV)}, 2023.

\bibitem[Shu et~al.(2022)Shu, Nie, Huang, Yu, Goldstein, Anandkumar, and
  Xiao]{shu2022test}
Manli Shu, Weili Nie, De-An Huang, Zhiding Yu, Tom Goldstein, Anima Anandkumar,
  and Chaowei Xiao.
\newblock Test-time prompt tuning for zero-shot generalization in
  vision-language models.
\newblock \emph{Advances in Neural Information Processing Systems},
  35:\penalty0 14274--14289, 2022.

\bibitem[Soomro et~al.(2012)Soomro, Zamir, and Shah]{soomro2012ucf101}
Khurram Soomro, Amir~Roshan Zamir, and Mubarak Shah.
\newblock Ucf101: A dataset of 101 human actions classes from videos in the
  wild.
\newblock \emph{arXiv preprint arXiv:1212.0402}, 2012.

\bibitem[Stokes(2015)]{stokes2015activity}
Mark~G Stokes.
\newblock ‘activity-silent’working memory in prefrontal cortex: a dynamic
  coding framework.
\newblock \emph{Trends in cognitive sciences}, 19\penalty0 (7):\penalty0
  394--405, 2015.

\bibitem[Sukhbaatar et~al.(2015)Sukhbaatar, Weston, Fergus,
  et~al.]{sukhbaatar2015end}
Sainbayar Sukhbaatar, Jason Weston, Rob Fergus, et~al.
\newblock End-to-end memory networks.
\newblock \emph{Advances in neural information processing systems}, 28, 2015.

\bibitem[Sun et~al.(2023)Sun, Wu, Zhang, Yong, and Zhang]{sun2023improving}
Lingchen Sun, Rongyuan Wu, Zhengqiang Zhang, Hongwei Yong, and Lei Zhang.
\newblock Improving the stability of diffusion models for content consistent
  super-resolution.
\newblock \emph{arXiv preprint arXiv:2401.00877}, 2023.

\bibitem[Udandarao et~al.(2023)Udandarao, Gupta, and Albanie]{udandarao2023sus}
Vishaal Udandarao, Ankush Gupta, and Samuel Albanie.
\newblock Sus-x: Training-free name-only transfer of vision-language models.
\newblock In \emph{Proceedings of the IEEE/CVF International Conference on
  Computer Vision}, pages 2725--2736, 2023.

\bibitem[Wang et~al.(2019)Wang, Ge, Lipton, and Xing]{wang2019learning}
Haohan Wang, Songwei Ge, Zachary Lipton, and Eric~P Xing.
\newblock Learning robust global representations by penalizing local predictive
  power.
\newblock \emph{Advances in Neural Information Processing Systems}, 32, 2019.

\bibitem[Weston et~al.(2014)Weston, Chopra, and Bordes]{weston2014memory}
Jason Weston, Sumit Chopra, and Antoine Bordes.
\newblock Memory networks.
\newblock \emph{arXiv preprint arXiv:1410.3916}, 2014.

\bibitem[Wortsman et~al.(2022)Wortsman, Ilharco, Kim, Li, Kornblith, Roelofs,
  Lopes, Hajishirzi, Farhadi, Namkoong, et~al.]{wortsman2022robust}
Mitchell Wortsman, Gabriel Ilharco, Jong~Wook Kim, Mike Li, Simon Kornblith,
  Rebecca Roelofs, Raphael~Gontijo Lopes, Hannaneh Hajishirzi, Ali Farhadi,
  Hongseok Namkoong, et~al.
\newblock Robust fine-tuning of zero-shot models.
\newblock In \emph{Proceedings of the IEEE/CVF Conference on Computer Vision
  and Pattern Recognition}, pages 7959--7971, 2022.

\bibitem[Wu et~al.(2024)Wu, Yang, Sun, Zhang, Li, and Zhang]{wu2024seesr}
Rongyuan Wu, Tao Yang, Lingchen Sun, Zhengqiang Zhang, Shuai Li, and Lei Zhang.
\newblock Seesr: Towards semantics-aware real-world image super-resolution.
\newblock In \emph{Proceedings of the IEEE/CVF conference on computer vision
  and pattern recognition}, 2024.

\bibitem[Xiao et~al.(2010)Xiao, Hays, Ehinger, Oliva, and
  Torralba]{xiao2010sun}
Jianxiong Xiao, James Hays, Krista~A Ehinger, Aude Oliva, and Antonio Torralba.
\newblock Sun database: Large-scale scene recognition from abbey to zoo.
\newblock In \emph{2010 IEEE computer society conference on computer vision and
  pattern recognition}, pages 3485--3492. IEEE, 2010.

\bibitem[Xie et~al.(2021)Xie, Xiong, Liu, Yao, and Shao]{xie2021few}
Guo-Sen Xie, Huan Xiong, Jie Liu, Yazhou Yao, and Ling Shao.
\newblock Few-shot semantic segmentation with cyclic memory network.
\newblock In \emph{Proceedings of the IEEE/CVF International Conference on
  Computer Vision}, pages 7293--7302, 2021.

\bibitem[Xing et~al.(2023)Xing, Wu, Cheng, Zhang, Liang, Wang, and
  Zhang]{xing2023dual}
Yinghui Xing, Qirui Wu, De Cheng, Shizhou Zhang, Guoqiang Liang, Peng Wang, and
  Yanning Zhang.
\newblock Dual modality prompt tuning for vision-language pre-trained model.
\newblock \emph{IEEE Transactions on Multimedia}, 2023.

\bibitem[Yang et~al.(2022)Yang, Duan, Tran, Xu, Chanda, Chen, Zeng, Chilimbi,
  and Huang]{yang2022vision}
Jinyu Yang, Jiali Duan, Son Tran, Yi Xu, Sampath Chanda, Liqun Chen, Belinda
  Zeng, Trishul Chilimbi, and Junzhou Huang.
\newblock Vision-language pre-training with triple contrastive learning.
\newblock In \emph{Proceedings of the IEEE/CVF Conference on Computer Vision
  and Pattern Recognition}, pages 15671--15680, 2022.

\bibitem[Yu et~al.(2023)Yu, Lu, Jin, Chen, and Wang]{yu2023task}
Tao Yu, Zhihe Lu, Xin Jin, Zhibo Chen, and Xinchao Wang.
\newblock Task residual for tuning vision-language models.
\newblock In \emph{Proceedings of the IEEE/CVF Conference on Computer Vision
  and Pattern Recognition}, pages 10899--10909, 2023.

\bibitem[Zang et~al.(2022)Zang, Li, Zhou, Huang, and Loy]{zang2022unified}
Yuhang Zang, Wei Li, Kaiyang Zhou, Chen Huang, and Chen~Change Loy.
\newblock Unified vision and language prompt learning.
\newblock \emph{arXiv preprint arXiv:2210.07225}, 2022.

\bibitem[Zhang et~al.(2023{\natexlab{a}})Zhang, Su, Xu, and
  Jia]{zhang2023improving}
Haojie Zhang, Yongyi Su, Xun Xu, and Kui Jia.
\newblock Improving the generalization of segmentation foundation model under
  distribution shift via weakly supervised adaptation.
\newblock \emph{arXiv preprint arXiv:2312.03502}, 2023{\natexlab{a}}.

\bibitem[Zhang et~al.(2021)Zhang, Fang, Zhang, Gao, Li, Dai, Qiao, and
  Li]{zhang2021tip}
Renrui Zhang, Rongyao Fang, Wei Zhang, Peng Gao, Kunchang Li, Jifeng Dai, Yu
  Qiao, and Hongsheng Li.
\newblock Tip-adapter: Training-free clip-adapter for better vision-language
  modeling.
\newblock \emph{arXiv preprint arXiv:2111.03930}, 2021.

\bibitem[Zhang et~al.(2022{\natexlab{a}})Zhang, Guo, Zhang, Li, Miao, Cui,
  Qiao, Gao, and Li]{zhang2022pointclip}
Renrui Zhang, Ziyu Guo, Wei Zhang, Kunchang Li, Xupeng Miao, Bin Cui, Yu Qiao,
  Peng Gao, and Hongsheng Li.
\newblock Pointclip: Point cloud understanding by clip.
\newblock In \emph{Proceedings of the IEEE/CVF Conference on Computer Vision
  and Pattern Recognition}, pages 8552--8562, 2022{\natexlab{a}}.

\bibitem[Zhang et~al.(2023{\natexlab{b}})Zhang, Hu, Li, Huang, Deng, Qiao, Gao,
  and Li]{zhang2023prompt}
Renrui Zhang, Xiangfei Hu, Bohao Li, Siyuan Huang, Hanqiu Deng, Yu Qiao, Peng
  Gao, and Hongsheng Li.
\newblock Prompt, generate, then cache: Cascade of foundation models makes
  strong few-shot learners.
\newblock In \emph{Proceedings of the IEEE/CVF Conference on Computer Vision
  and Pattern Recognition}, pages 15211--15222, 2023{\natexlab{b}}.

\bibitem[Zhang et~al.(2020)Zhang, Deng, Tang, Zhang, and
  Jia]{zhang2020unsupervised}
Yabin Zhang, Bin Deng, Hui Tang, Lei Zhang, and Kui Jia.
\newblock Unsupervised multi-class domain adaptation: Theory, algorithms, and
  practice.
\newblock \emph{IEEE Transactions on Pattern Analysis and Machine
  Intelligence}, 44\penalty0 (5):\penalty0 2775--2792, 2020.

\bibitem[Zhang et~al.(2022{\natexlab{b}})Zhang, Li, Li, Jia, and
  Zhang]{zhang2022exact}
Yabin Zhang, Minghan Li, Ruihuang Li, Kui Jia, and Lei Zhang.
\newblock Exact feature distribution matching for arbitrary style transfer and
  domain generalization.
\newblock In \emph{Proceedings of the IEEE/CVF conference on computer vision
  and pattern recognition}, pages 8035--8045, 2022{\natexlab{b}}.

\bibitem[Zhou et~al.(2022{\natexlab{a}})Zhou, Yang, Loy, and
  Liu]{zhou2022conditional}
Kaiyang Zhou, Jingkang Yang, Chen~Change Loy, and Ziwei Liu.
\newblock Conditional prompt learning for vision-language models.
\newblock In \emph{Proceedings of the IEEE/CVF Conference on Computer Vision
  and Pattern Recognition}, pages 16816--16825, 2022{\natexlab{a}}.

\bibitem[Zhou et~al.(2022{\natexlab{b}})Zhou, Yang, Loy, and
  Liu]{zhou2022learning}
Kaiyang Zhou, Jingkang Yang, Chen~Change Loy, and Ziwei Liu.
\newblock Learning to prompt for vision-language models.
\newblock \emph{International Journal of Computer Vision}, 130\penalty0
  (9):\penalty0 2337--2348, 2022{\natexlab{b}}.

\bibitem[Zhou et~al.(2023)Zhou, Ren, Li, Zabih, and Lim]{zhou2023distribution}
Yifei Zhou, Juntao Ren, Fengyu Li, Ramin Zabih, and Ser-Nam Lim.
\newblock Distribution normalization: An" effortless" test-time augmentation
  for contrastively learned visual-language models.
\newblock \emph{arXiv preprint arXiv:2302.11084}, 2023.

\bibitem[Zhu et~al.(2023{\natexlab{a}})Zhu, Niu, Han, Wu, and
  Zhang]{zhu2023prompt}
Beier Zhu, Yulei Niu, Yucheng Han, Yue Wu, and Hanwang Zhang.
\newblock Prompt-aligned gradient for prompt tuning.
\newblock In \emph{Proceedings of the IEEE/CVF International Conference on
  Computer Vision}, pages 15659--15669, 2023{\natexlab{a}}.

\bibitem[Zhu et~al.(2023{\natexlab{b}})Zhu, Zhang, He, Zhou, Wang, Zhao, and
  Gao]{zhu2023not}
Xiangyang Zhu, Renrui Zhang, Bowei He, Aojun Zhou, Dong Wang, Bin Zhao, and
  Peng Gao.
\newblock Not all features matter: Enhancing few-shot clip with adaptive prior
  refinement.
\newblock \emph{arXiv preprint arXiv:2304.01195}, 2023{\natexlab{b}}.

\end{thebibliography}
}

 \clearpage
\setcounter{page}{1}
\maketitlesupplementary
\appendix
\renewcommand{\thetable}{A\arabic{table}}
\renewcommand{\thefigure}{A\arabic{figure}}
\numberwithin{equation}{section}

\begin{figure*}[htb]
	\centering
	\begin{subfigure}{0.33\textwidth}
		\includegraphics[width=\textwidth]{images/results_path_resnet_training_free/resnet_tf_Mean.pdf}
	\end{subfigure}
	\hfill
	\begin{subfigure}{0.33\textwidth}
		\includegraphics[width=\textwidth]{images/results_path_resnet_training_free/resnet_tf_ImageNet.pdf}
	\end{subfigure}
	\hfill
	\begin{subfigure}{0.33\textwidth}
		\includegraphics[width=\textwidth]{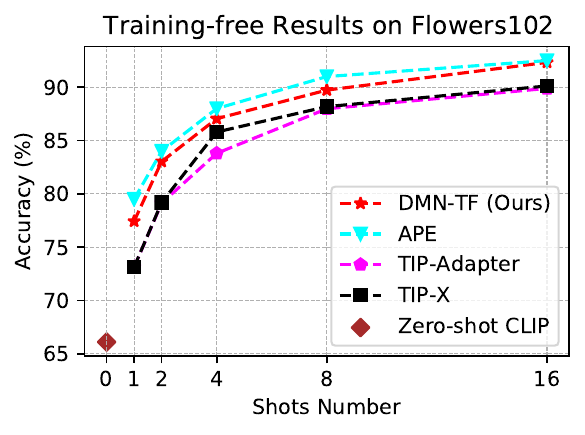}
	\end{subfigure}
	\hfill
	\begin{subfigure}{0.33\textwidth}
		\includegraphics[width=\textwidth]{images/results_path_resnet_training_free/resnet_tf_DTD.pdf}
	\end{subfigure}
	\begin{subfigure}{0.33\textwidth}
		\includegraphics[width=\textwidth]{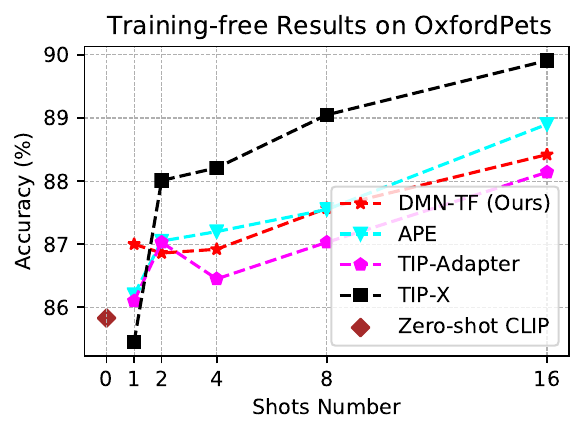}
	\end{subfigure}
	\hfill
	\begin{subfigure}{0.33\textwidth}
		\includegraphics[width=\textwidth]{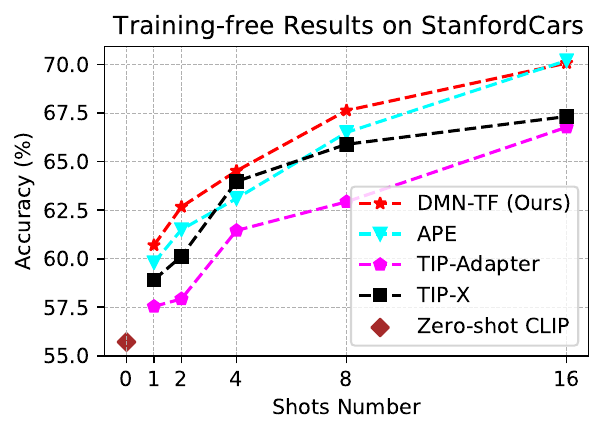}
	\end{subfigure}
	\hfill
	\begin{subfigure}{0.33\textwidth}
		\includegraphics[width=\textwidth]{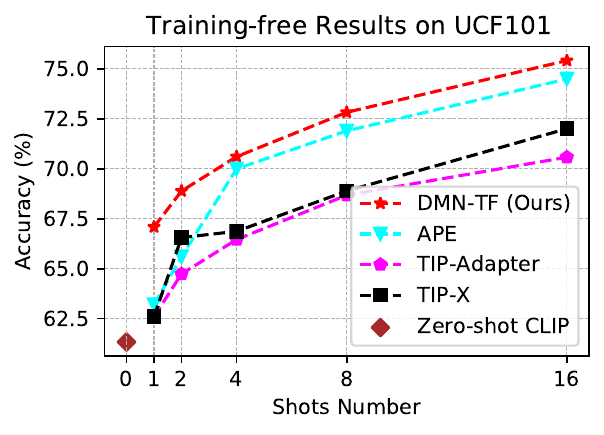}
	\end{subfigure}
	\hfill
	\begin{subfigure}{0.33\textwidth}
		\includegraphics[width=\textwidth]{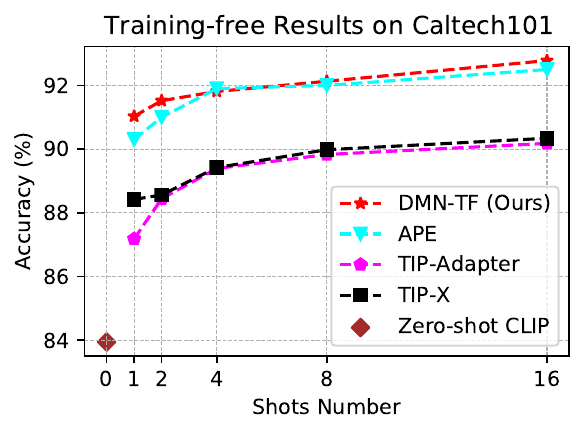}
	\end{subfigure}
	\begin{subfigure}{0.33\textwidth}
		\includegraphics[width=\textwidth]{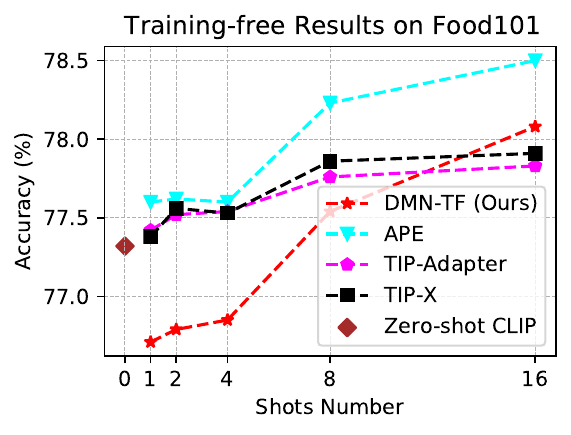}
	\end{subfigure}
	\hfill
	\begin{subfigure}{0.33\textwidth}
		\includegraphics[width=\textwidth]{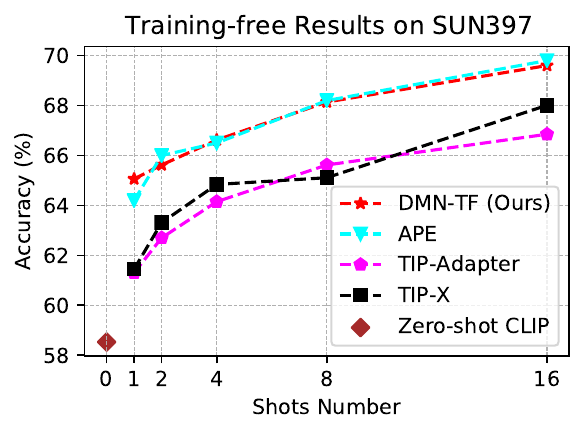}
	\end{subfigure}
	\hfill
	\begin{subfigure}{0.33\textwidth}
		\includegraphics[width=\textwidth]{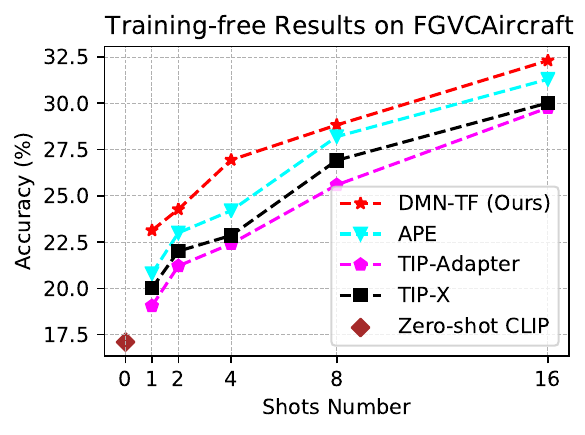}
	\end{subfigure}
	\hfill
	\begin{subfigure}{0.33\textwidth}
		\includegraphics[width=\textwidth]{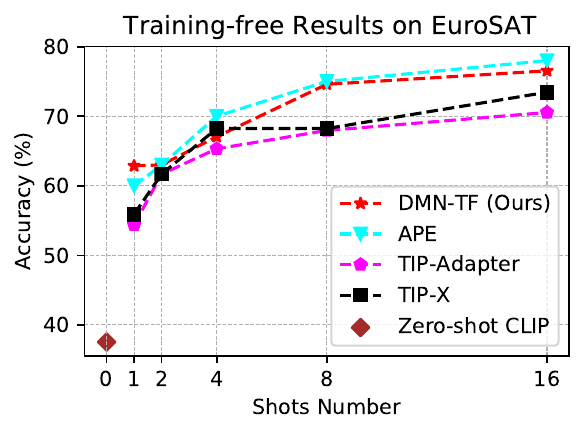}
	\end{subfigure}
	\caption{Training-free few-shot results of our DMN-TF and other methods on $11$ classification datasets with the ResNet50 backbone. } \label{Fig:full_tf_res50}
\end{figure*}

\begin{figure*}[ht]
	\centering
	\begin{subfigure}{0.33\textwidth}
		\includegraphics[width=\textwidth]{images/results_path_vit_training_required/vit_tr_Mean.pdf}
	\end{subfigure}
	\hfill
	\begin{subfigure}{0.33\textwidth}
		\includegraphics[width=\textwidth]{images/results_path_vit_training_required/vit_tr_ImageNet.pdf}
	\end{subfigure}
	\hfill
	\begin{subfigure}{0.33\textwidth}
		\includegraphics[width=\textwidth]{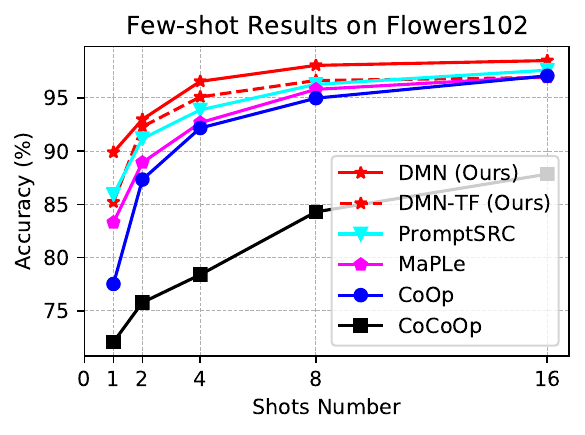}
	\end{subfigure}
	\hfill
	\begin{subfigure}{0.33\textwidth}
		\includegraphics[width=\textwidth]{images/results_path_vit_training_required/vit_tr_DTD.pdf}
	\end{subfigure}
	\begin{subfigure}{0.33\textwidth}
		\includegraphics[width=\textwidth]{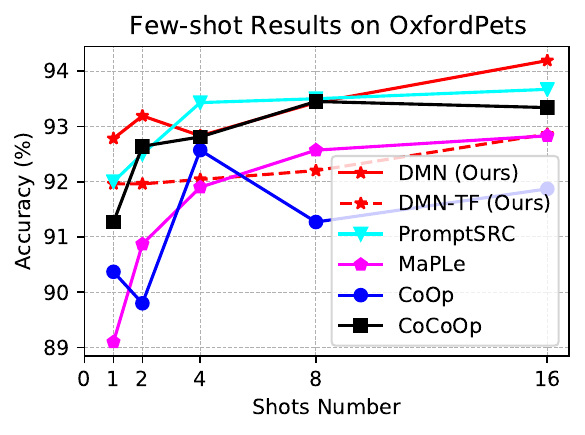}
	\end{subfigure}
	\hfill
	\begin{subfigure}{0.33\textwidth}
		\includegraphics[width=\textwidth]{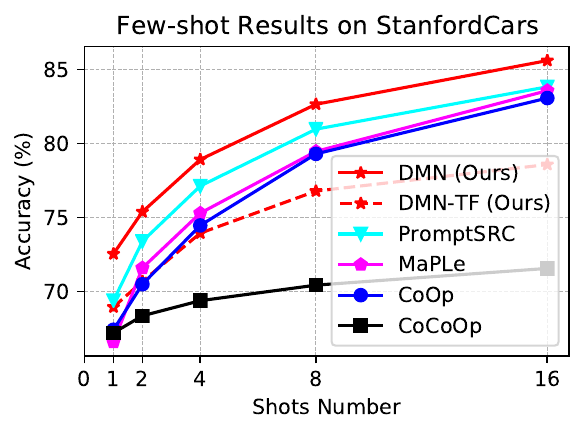}
	\end{subfigure}
	\hfill
	\begin{subfigure}{0.33\textwidth}
		\includegraphics[width=\textwidth]{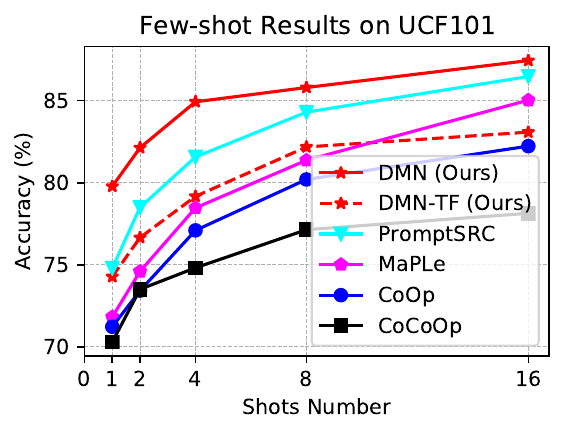}
	\end{subfigure}
	\hfill
	\begin{subfigure}{0.33\textwidth}
		\includegraphics[width=\textwidth]{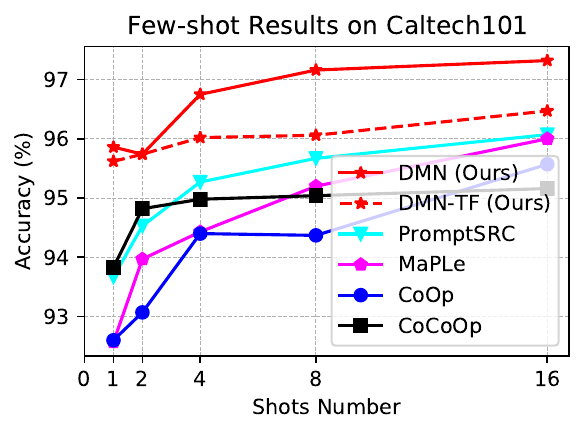}
	\end{subfigure}
	\begin{subfigure}{0.33\textwidth}
		\includegraphics[width=\textwidth]{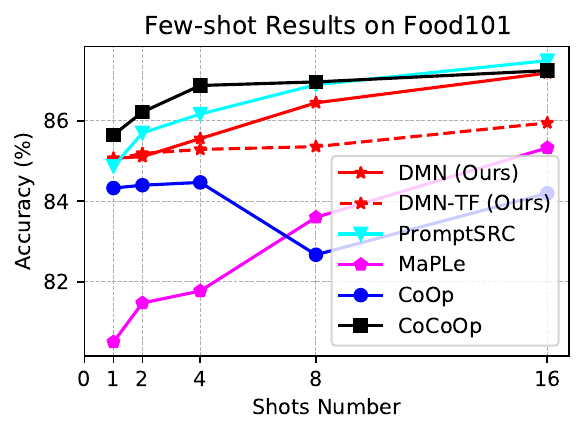}
	\end{subfigure}
	\hfill
	\begin{subfigure}{0.33\textwidth}
		\includegraphics[width=\textwidth]{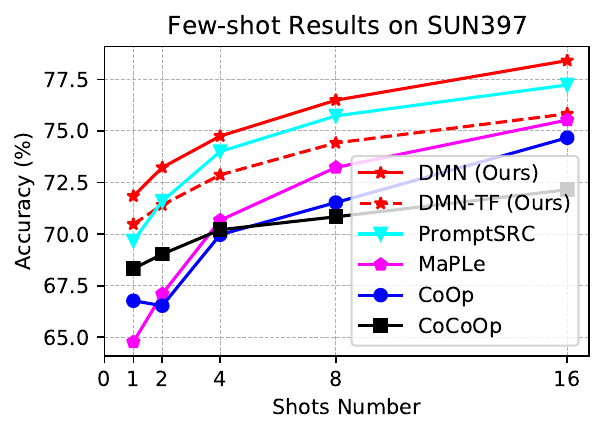}
	\end{subfigure}
	\hfill
	\begin{subfigure}{0.33\textwidth}
		\includegraphics[width=\textwidth]{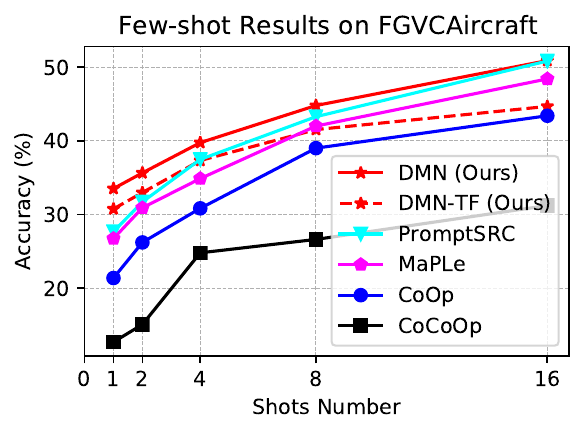}
	\end{subfigure}
	\hfill
	\begin{subfigure}{0.33\textwidth}
		\includegraphics[width=\textwidth]{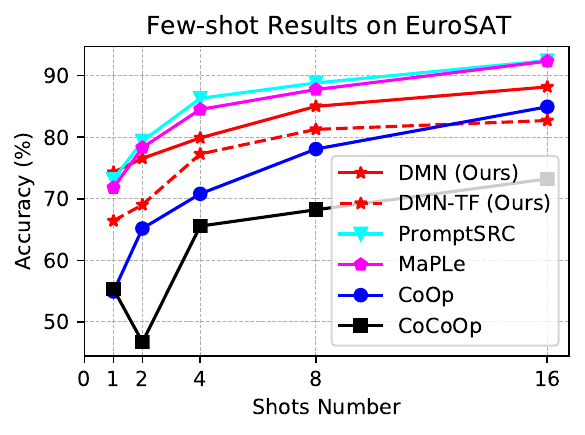}
	\end{subfigure}
	\caption{Few-shot results of our DMN and other methods on $11$ classification datasets with the VITB/16 backbone. }\label{Fig:full_tr_vit}
\end{figure*}

\begin{figure*}[ht]
	\centering
	\begin{subfigure}{0.33\textwidth}
		\includegraphics[width=\textwidth]{images/results_path_resnet_training_required/resnet_tr_Mean.pdf}
	\end{subfigure}
	\hfill
	\begin{subfigure}{0.33\textwidth}
		\includegraphics[width=\textwidth]{images/results_path_resnet_training_required/resnet_tr_ImageNet.pdf}
	\end{subfigure}
	\hfill
	\begin{subfigure}{0.33\textwidth}
		\includegraphics[width=\textwidth]{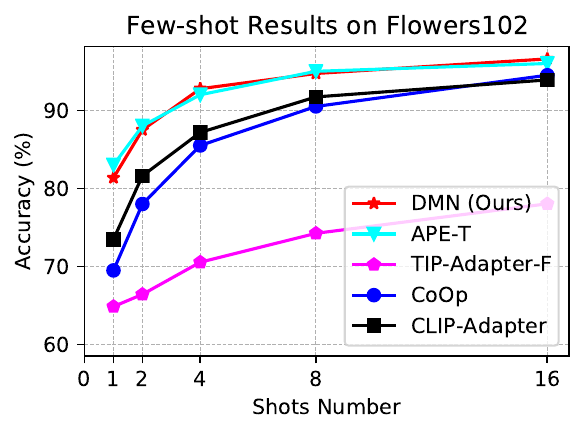}
	\end{subfigure}
	\hfill
	\begin{subfigure}{0.33\textwidth}
		\includegraphics[width=\textwidth]{images/results_path_resnet_training_required/resnet_tr_DTD.pdf}
	\end{subfigure}
	\begin{subfigure}{0.33\textwidth}
		\includegraphics[width=\textwidth]{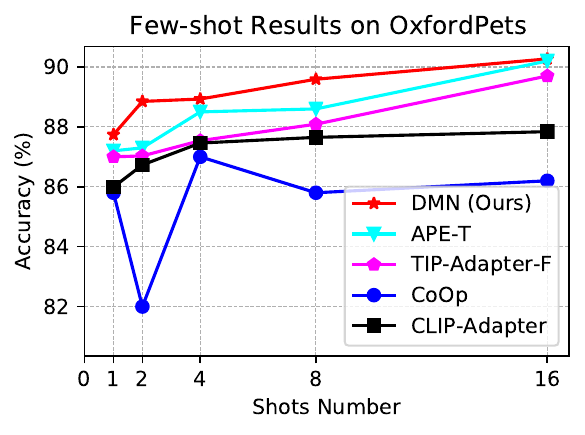}
	\end{subfigure}
	\hfill
	\begin{subfigure}{0.33\textwidth}
		\includegraphics[width=\textwidth]{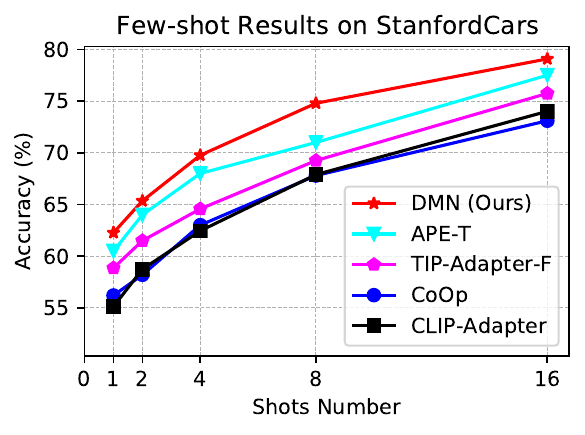}
	\end{subfigure}
	\hfill
	\begin{subfigure}{0.33\textwidth}
		\includegraphics[width=\textwidth]{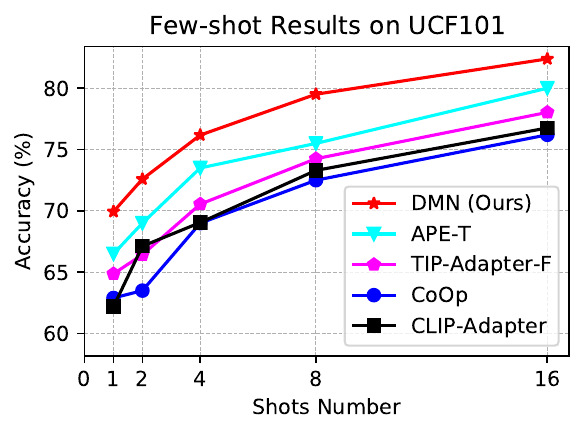}
	\end{subfigure}
	\hfill
	\begin{subfigure}{0.33\textwidth}
		\includegraphics[width=\textwidth]{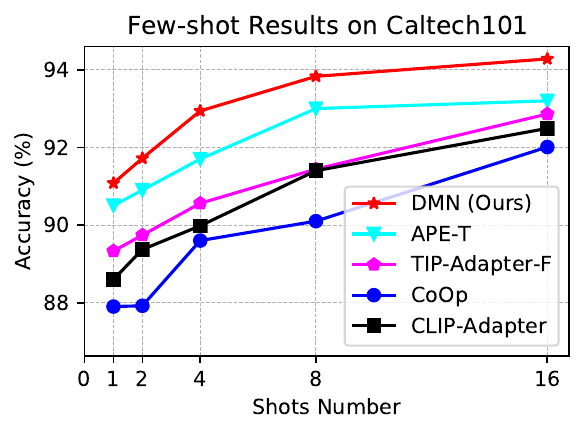}
	\end{subfigure}
	\begin{subfigure}{0.33\textwidth}
		\includegraphics[width=\textwidth]{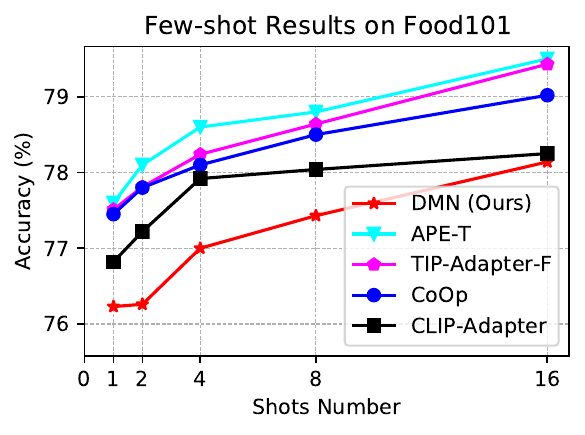}
	\end{subfigure}
	\hfill
	\begin{subfigure}{0.33\textwidth}
		\includegraphics[width=\textwidth]{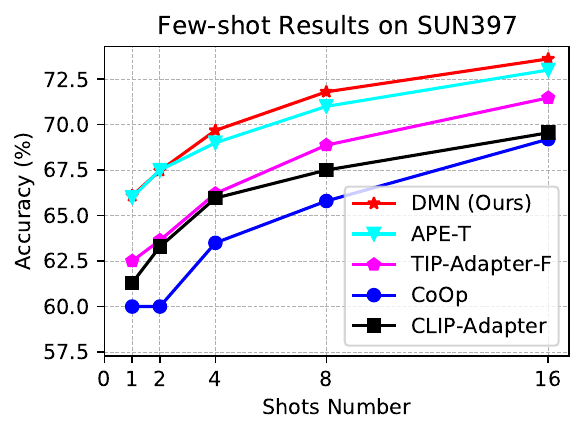}
	\end{subfigure}
	\hfill
	\begin{subfigure}{0.33\textwidth}
		\includegraphics[width=\textwidth]{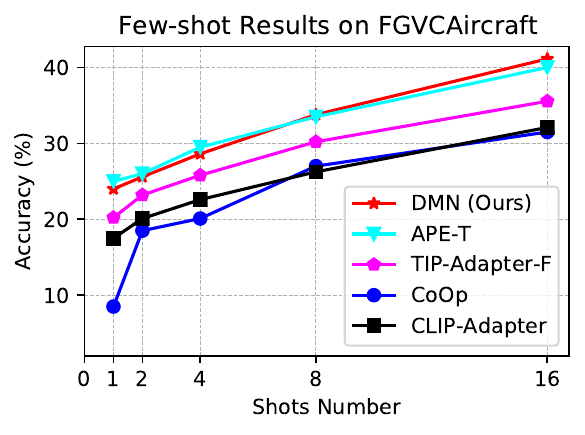}
	\end{subfigure}
	\hfill
	\begin{subfigure}{0.33\textwidth}
		\includegraphics[width=\textwidth]{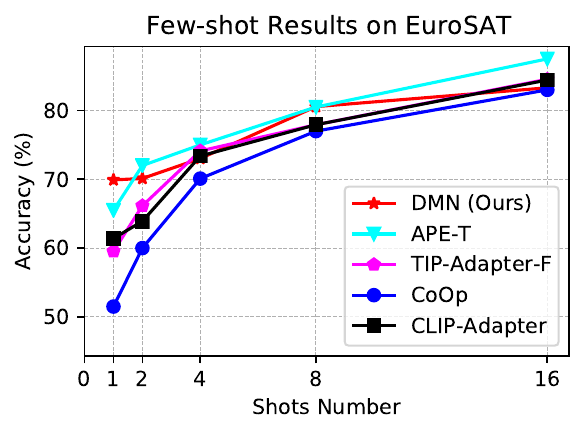}
	\end{subfigure}
	\caption{Few-shot results of our DMN and other methods on $11$ classification datasets with the ResNet50 backbone. } \label{Fig:full_tr_res50}
\end{figure*}

\begin{table*} [b]
	\centering
	\begin{tabular}{l|l|ccccccccccc}
		\toprule
		Settings & Items & ImageNet & Flower & DTD & Pets & Cars & UCF & Caltech & Food & SUN & Aircraft & EuroSAT    \\
		\midrule
		\multirow{2}{*}{1shot} 
		& $\alpha_2$ & 1.0 & 0.3 & 0.3   & 1.0 & 1.0    & 100 & 0.3   & 0.3 &3.0 & 0.3   & 1.0 \\
		& $\alpha_3$ & 0.1 & 1.0 & 0.03 & 0.3 & 0.001 & 3.0 & 0.001&0.1 &1.0 & 0.001&0.1 \\
		\midrule
		\multirow{2}{*}{2shot} 
		& $\alpha_2$ & 1.0 & 0.3 & 1.0 & 1.0     & 1.0  & 0.3  & 0.3 & 0.3    & 1.0  & 3.0 & 1.0\\
		& $\alpha_3$ & 0.3 & 1.0 & 1.0 & 0.001 & 0.03 & 0.03& 0.3 & 0.001 & 0.3 & 0.3 & 1.0 \\
		\midrule
		\multirow{2}{*}{4shot} 
		& $\alpha_2$ & 1.0 & 0.3 & 0.3 & 1.0  & 1.0  & 3.0 & 1.0 & 0.3 & 0.3 & 1.0 & 1.0\\
		& $\alpha_3$ & 0.3 & 1.0 & 0.3 & 0.03& 0.03 & 3.0& 0.3 & 0.1 & 0.3 & 1.0 & 1.0 \\
		\midrule
		\multirow{2}{*}{8shot} & $\alpha_2$ & 1.0 & 1.0 & 0.1 & 1.0 & 3.0     & 1.0 & 0.3 & 0.3  & 0.3     & 3.0 & 0.3  \\
		& $\alpha_3$ & 0.3 & 1.0 & 0.1 & 0.3 & 0.001 & 0.3 & 0.3 & 0.03 & 0.001 & 3.0 & 1.0 \\
		\midrule
		\multirow{2}{*}{16shot} 
		& $\alpha_2$ &  1.0 & 3.0 & 0.3  & 1.0  &3.0    & 1.0 & 1.0  & 0.3  & 1.0   & 0.3 & 0.1\\
		& $\alpha_3$ &  1.0 & 1.0 & 0.03 & 0.03&0.001& 0.1& 0.001& 0.03& 0.01& 3.0 & 1.0\\
		\bottomrule
	\end{tabular}
	\vspace{-0.2cm}
	\caption{Searched optimal classifier weights of DMN for different task settings and datasets with the VITB/16 backbone. } \label{Tab:searched_weights}
\end{table*}

The following materials are provided in this supplementary file:
\begin{itemize}
        \item Discussion with Test-time Adaptation.
	\item Full results of few-shot classification (\cf Section \ref{Sec:performance_results} in the main paper).
	\item More analyses (\cf Section \ref{Subsec:ablation_analyses} in the main paper).
\end{itemize}

\section{Discussion with Test-time Adaptation (TTA)}

Our approach, especially the DMN-ZS variant, shares some high-level ideas with TTA methods \cite{shu2022test,feng2023diverse} by updating the model (\eg, memory) at test time.
However, there are some key distinctions.
First, unlike \cite{shu2022test,feng2023diverse}, we leverage all historical test samples (not just the current one), improving the results by 3.77\% (cf. Tab. \ref{Tab:zero_shot}). 
Second, we avoid test-time optimization, maintaining fast test speed (cf. Tab. \ref{Tab:computation_efficiency}).
Third, we integrate the utilization of test and training data via flexible memory networks, extending the applicability, \eg, few-shot classification (cf. Tab. \ref{Tab:summary_clip_adapt}).

\section{Full Results of Few-shot Classification}
The full results of training-free few-shot classification and traditional few-shot classification are presented in Figures \ref{Fig:full_tf_res50}, \ref{Fig:full_tr_vit}, and \ref{Fig:full_tr_res50}.
Similar to the observations in the main paper, our DMN consistently surpasses competing approaches in terms of average accuracy across 11 datasets, maintaining superiority with different backbone architectures and varying numbers of training samples. 
On individual datasets, although our method occasionally lags behind other state-of-the-art methods in certain settings (\eg, the Food101 dataset), it achieves consistent gains on the acknowledged ImageNet dataset, affirming its effectiveness.

\section{More Analyses}
\begin{figure*}
	\centering
	\includegraphics[width=0.4\linewidth]{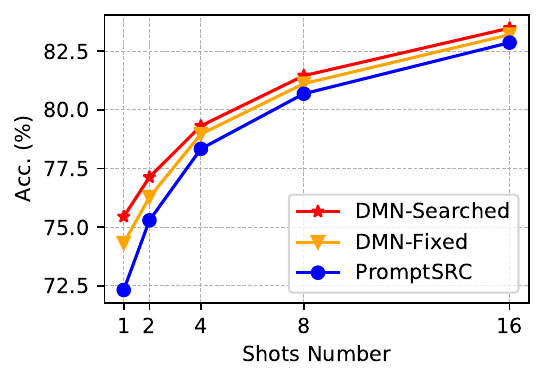}
	\caption{Average results of DMN on 11 datasets with the VITB/16 backbone.  DMN-Searched and DMN-Fixed represent results with searched and fixed classifier weights, respectively. We also provide results of the recent PromptSRC method for reference. } \label{Fig:searched_fixed}
\end{figure*}

\textbf{Classifier Weights.}
We fix $\alpha_1 =1.0$ in Eq. (\ref{Equ:dmn_three_pred}) and search for the optimal $\alpha_2$ and $\alpha_3$ for each downstream task.
The discrete search space for $\alpha_2$ and $\alpha_3$  is $\{ 0.001, 0.003, 0.01, 0.03, 0.1, 0.3, 1, 3, 10, 30, 100, 300 \}$.
The searched optimal classifier weights are shown in Tab. \ref{Tab:searched_weights}.
We can observe that the value of $\alpha_2$ is typically larger than that of $\alpha_3$, highlighting the importance of historical test knowledge. 
We also find that fixing $\alpha_2 =1.0$ and $\alpha_3 =0.3$ can generally lead to good results in different task settings, as presented in Fig. \ref{Fig:searched_fixed}.

\begin{figure*}
	\centering
	\includegraphics[width=0.4\linewidth]{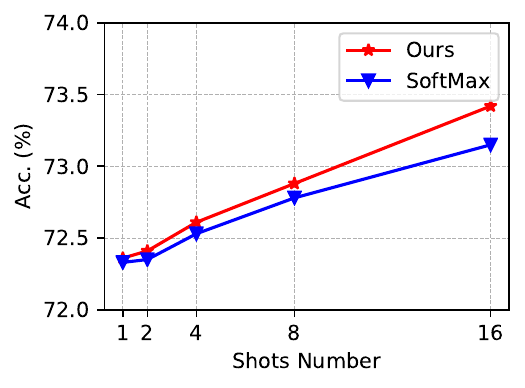}
	\caption{Results of DMN-TF with different non-linear functions on ImageNet dataset, where the VITB/16 backbone is adopted.} \label{Fig:softmax}
\end{figure*}

\textbf{Non-linear Function $\varphi(\cdot)$.}
We compare the adopted non-linear function $\varphi(x) = \exp(-\beta (1-x))$ with the popular SoftMax function, \ie, $\mathrm{SoftMax(\beta x)}$. 
We also search for the optimal $\beta$ for the SoftMax function.
As shown in Fig. \ref{Fig:softmax}, our strategy typically outperforms the popular SoftMax function. 
The possible reason for this could be that the output of SoftMax is influenced by both the value of a single element and its relative size compared to other elements. 
Therefore, the output of SoftMax is directly related to the memory length. 
In our method, the effective memory length varies due to the different shot numbers and the online update of dynamic memory, which may affect the usage of SoftMax function. 
In contrast, the output of our adopted $\varphi(\cdot)$ only depends on the value of a single element, making it more suitable for our task setting.

\textbf{Test Data Order.}
By managing test data order with random seeds, we observed slight performance variations. For instance, DMN-ZS scored 72.25$\pm$0.21\% on ImageNet over 3 random runs.

%\begin{table}[htb]
%	\centering
%	\begin{tabular}{lcccccccccc}
%		\hline
%		DMN-TF      & 0.1    & 0.3     &  1.0    & 3.0     & 10     & 30      & 100   & 300 & used\\
%		SoftMax      & 71.98 & 71.99 &  72.07 & 72.27 & 72.72& 72.43  & 71.71 & 71.32 & 72.54\\
%		\hline
%	\end{tabular}
%	\caption{Softmax Non-linear functions. 3/3}
%\end{table}

%\begin{table}[H]
%	\centering
%	\begin{tabular}{lccccccccc}
%		\hline
%		DMN-TF      & 0.1    & 0.3     &  1.0    & 3.0     & 10     & 30      & 100   & 300 & used\\
%		SoftMax-16shot & 72.07      &  72.11  & 72.26 & 72.88   & \textbf{73.05} & 72.79 & 72.27 &&73.42 \\
%		SoftMax-4shot   & &&71.99 & \textbf{72.53} & 72.33 & 72.07 & 71.99 &&72.61  \\
%		SoftMax-0shot   & &71.61 & 71.88 & \textbf{72.25} & 71.92 & & & & 72.25\\
%		\hline
%	\end{tabular}
%	\caption{Softmax Non-linear functions. SMN-TF. 3/32}
%\end{table}

%\section{More Results Beyond CLIP.}
%Other Vision-language models.

%\section{Rationale}
%\label{sec:rationale}
%% 
%Having the supplementary compiled together with the main paper means that:
%% 
%\begin{itemize}
%	\item The supplementary can back-reference sections of the main paper, for example, we can refer to \cref{sec:intro};
%	\item The main paper can forward reference sub-sections within the supplementary explicitly (e.g. referring to a particular experiment); 
%	\item When submitted to arXiv, the supplementary will already included at the end of the paper.
%\end{itemize}
%% 
%To split the supplementary pages from the main paper, you can use \href{https://support.apple.com/en-ca/guide/preview/prvw11793/mac#:~:text=Delete%20a%20page%20from%20a,or%20choose%20Edit%20%3E%20Delete).}{Preview (on macOS)}, \href{https://www.adobe.com/acrobat/how-to/delete-pages-from-pdf.html#:~:text=Choose%20%E2%80%9CTools%E2%80%9D%20%3E%20%E2%80%9COrganize,or%20pages%20from%20the%20file.}{Adobe Acrobat} (on all OSs), as well as \href{https://superuser.com/questions/517986/is-it-possible-to-delete-some-pages-of-a-pdf-document}{command line tools}.

\end{document}